\documentclass{article}

\usepackage[main, final]{neurips_2026}

\usepackage[utf8]{inputenc} 
\usepackage[T1]{fontenc}    
\usepackage{hyperref}       
\usepackage{url}            
\usepackage{booktabs}       
\usepackage{amsfonts}       
\usepackage{nicefrac}       
\usepackage{microtype}      
\usepackage{xcolor}         

\usepackage{graphicx}
\usepackage{subcaption}
\usepackage[table]{xcolor}
\definecolor{tablehighlight}{gray}{0.92}
\definecolor{mylightblue}{RGB}{235, 235, 255}
\usepackage{amsmath}
\usepackage{amssymb}
\usepackage{mathtools}
\usepackage{amsthm}
\usepackage{multirow}

\theoremstyle{plain}

\theoremstyle{definition}

\theoremstyle{remark}

\usepackage{float}
\def\ourmethod{FreshMem}
\usepackage[capitalize,noabbrev]{cleveref}
\title{\textit{FreshMem}: Brain-Inspired Frequency-Space Hybrid Memory for Streaming Video Understanding}

\author{
  \textbf{Kangcong Li}$^{1}$,\quad
  \textbf{Peng Ye}$^{2,3*}$,\quad
  \textbf{Lin Zhang}$^{1}$,\quad  \textbf{Chao Wang}$^{1}$, \\
  \textbf{Huafeng Qin}$^{4}$, \quad
  \textbf{Jiayuan Fan}$^{1}$\thanks{Corresponding authors: \texttt{jyfan@fudan.edu.cn}.},\quad
  \textbf{Tao Chen}$^{1,5}$ 
  \\
  $^{1}$Fudan University \quad
  $^{2}$Shanghai AI Lab \quad
  $^{3}$CUHK \\
  $^{4}$Chongqing Technology and Business University \\
  $^{5}$Shanghai Innovation Institute
}

\begin{document}

\maketitle

\begin{abstract}
  Transitioning Multimodal Large Language Models (MLLMs) from offline to online streaming video understanding is essential for continuous perception. 
  However, existing methods lack flexible adaptivity, leading to irreversible detail loss and context fragmentation. 
  To resolve this, we propose \textbf{\textit{FreshMem}}, a \textbf{Fre}quency-\textbf{S}pace \textbf{H}ybrid \textbf{Mem}ory network inspired by the brain's logarithmic perception and memory consolidation. 
  FreshMem reconciles short-term fidelity with long-term coherence through two synergistic modules: Multi-scale Frequency Memory (MFM), which projects overflowing frames into representative frequency coefficients, complementing by residual details to reconstruct a global historical “gist”;
  and Space Thumbnail Memory (STM), which discretizes the continuous stream into episodic clusters by 
  employing an adaptive compression strategy to distill them into high-density space thumbnails.
  Extensive experiments show that FreshMem significantly boosts the Qwen2-VL baseline, yielding gains of 5.20\%, 4.52\%, and 2.34\% on StreamingBench, OV-Bench, and OVO-Bench, respectively.
  Besides its plug-and-play design, following low-cost component-wise fine-tuning, FreshMem outperforms current fully fine-tuned methods and achieves a state-of-the-art performance of 82.31\% on StreamingBench, exceeding the baseline by 13.31\%, offering a highly efficient paradigm for long-horizon streaming video understanding.
\end{abstract}
\section{Introduction}
\label{Introduction}

Multimodal Large Language Models (MLLMs)~\cite{yang2023dawn,achiam2023gpt,wang2024qwen2vl,chen2024internvl2} have witnessed remarkable progress in long video understanding~\cite{li2024llamavid,zhang2024long,shen2024longvu}, demonstrating sophisticated comprehension and reasoning capabilities across diverse tasks.
However, most existing methods predominantly operate under an offline paradigm~\cite{song2024moviechat,MA-LMM}, analyzing video sequences as static, complete entities with a priori access to the entire temporal context. 
As MLLMs gravitate toward more dynamic and interactive environments, there is an urgent need to transition from such post-hoc analysis to online streaming video understanding~\cite{chen2024videollmonline,huang2025ovbench}. This shift represents a fundamental evolution toward continuous perception, requiring models to reason over information as it unfolds under strict causal constraints. 
Thus, a key challenge emerges in maintaining a delicate balance between long-term information preservation and the seamless integration of novel, incoming frames, which is vital for MLLMs to achieve robust streaming video understanding.

Existing works have primarily evolved along two trajectories: input-level modulation~\cite{xu2025streamingvlm,chatterjee2025streaming}, which manages data influx by merging or sparsifying frames, and memory-augmented architectures, which utilize mechanisms like KV cache eviction~\cite{di2025rekv,ning2025livevlm}, global memory tokens condensation~\cite{qian2024streaming,zeng2025streamforest}, or external memory bank construction~\cite{xiong2025streaming,zhang2024flashvstream}. 
While these methods mitigate computational pressure, they often precipitate an irrevocable loss of fine-grained details or struggle to reconcile short-term temporal fidelity with long-term contextual coherence. 
A fundamental limitation remains in their uniform treatment of temporal dynamics, which relies on rigid heuristics or static downsampling. 
Consequently, current frameworks lack the flexibility to adaptively adjust representation density over time.
As the temporal horizon extends, insufficient memory consolidation leads to fragmented representations, compromising action continuity and semantic consistency, and ultimately limiting the model’s ability to sustain a coherent long-horizon narrative.


\begin{figure*}[t]
  \centering    \centerline{\includegraphics[width=1.0\textwidth]{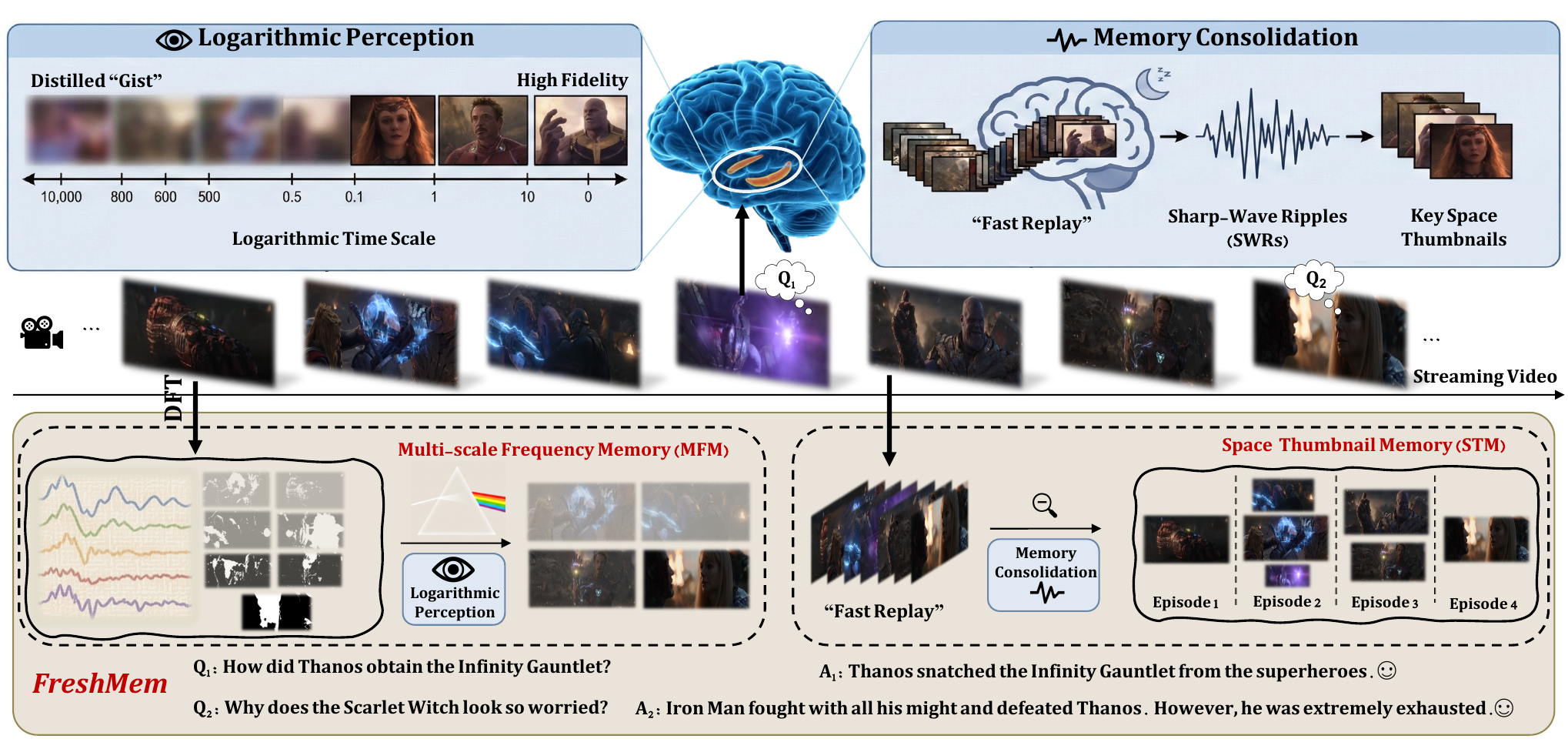}}
    \caption{\textbf{Overview of FreshMem and its biological foundations.} (Top-Left) Logarithmic Perception, where the brain maintains high fidelity for immediate stimuli while distilling the distant past into a semantic “gist”; (Top-Right) Memory Consolidation, where SWRs compress continuous experiences into key space thumbnails; (Bottom) FreshMem materializes these principles via two synergistic modules: MFM and STM. This hybrid design enables the model to effectively reconcile short-term fidelity with long-term coherence for streaming video understanding.}
    \label{fig1}
\end{figure*}

To address these structural deficiencies, we draw inspiration from the brain’s memory mechanisms, which effectively reconcile above bottleneck via two core mechanisms: logarithmic perception~\cite{law,logarithmicperception} for temporal organization and memory consolidation~\cite{memory,swr} for information refinement. 
Neuroscience research~\cite{logarithmicperception} indicates that human brain encodes time on a non-uniform, logarithmic scale, preserving immediate stimuli with high fidelity while progressively distilling the distant past into a semantic “gist”.
This suggests that a non-uniform temporal representation is fundamental for managing both local detail and global context. 
Complementing this, Neuroscience evidence~\cite{swr} reveals that the brain undergoes memory consolidation via mechanisms such as Sharp-Wave Ripples (SWRs), where extensive experiences are condensed into concise space thumbnails. 
This rapid replay and distillation process ensures the stability of long-term semantic structures.
By emulating these dual principles, a streaming model can theoretically enable a more adaptive approach to maintaining continuous representations over extended horizons.

To translate these biological insights into a deployable architecture, we propose \textbf{\textit{FreshMem}}, a \textbf{Fre}quency-\textbf{S}pace \textbf{H}ybrid \textbf{Mem}ory network. To embody logarithmic perception, we introduce the Multi-scale Frequency Memory (MFM) module. Rather than the uniform discarding prevalent in existing models, MFM utilizes Discrete Fourier Transforms (DFT)~\cite{li2026statisticalcharacteristicguideddenoisingrapid,fontana2026faarefficientfrequencyawaremultitask,gu2026fourieranglealignmentoriented,cai2026pfgnetfullyconvolutionalfrequencyguided} to project historical frames overflowing the sliding window~\cite{xu2025streamingvlm} into representative frequency coefficients. By using the superposition of these frequencies, MFM reconstructs the global “gist” of past context with minimal overhead, mitigating the raw data redundancy described earlier. Parallelly, the memory consolidation mechanism is materialized through a Space Thumbnail Memory (STM) module. By monitoring episodic boundaries via cosine similarity, STM discretizes the continuous stream into distinct episodic clusters. An adaptive compression strategy then transforms these clusters into high-density space thumbnails, utilizing a centroid-based fusion mechanism to merge episodes with high semantic affinity. 


To validate the effectiveness of the proposed FreshMem, we conduct extensive evaluations across both online streaming and offline video understanding benchmarks. 
In streaming scenarios, FreshMem delivers substantial performance gains over the competitive Qwen2-VL~\cite{wang2024qwen2vl} baseline, achieving absolute accuracy improvements of 5.20\% on StreamingBench~\cite{lin2024streamingbench}, 4.52\% on OV-Bench~\cite{huang2025ovbench}, and 2.34\% on OVO-Bench~\cite{niu2025ovobench}.
Notably, even as a training-free solution, FreshMem outperforms several fully fine-tuned counterparts, underscoring its superior capacity for handling continuous data streams. Furthermore, despite its plug-and-play design, FreshMem can be further enhanced through low-cost component-wise fine-tuning. This configuration achieves a state-of-the-art performance of 82.31\% on StreamingBench, representing a significant 13.31\% absolute improvement over the baseline and surpassing existing fully fine-tuned methods.
Beyond streaming tasks, FreshMem also maintains robust performance on traditional offline benchmarks, yielding consistent improvements on the long-video benchmark MLVU~\cite{zhou2025mlvu} and the short-video benchmark MVBench~\cite{li2024mvbench}. 
Overall, these results demonstrate that FreshMem effectively reconciles short-term fidelity with long-term coherence, offering a highly efficient and scalable alternative to computationally intensive paradigms.
The main contributions of our work are summarized as follows:

\textbf{Brain-inspired Paradigm for Streaming Video.}
We bridge the gap between biological memory mechanisms and streaming MLLMs by mimicking the human brain's logarithmic perception and memory consolidation. This novel perspective effectively addresses the structural dilemma between maintaining short-term fidelity and ensuring long-term semantic coherence in continuous video streams.

\textbf{Frequency-Space Hybrid Memory Network.}
We propose FreshMem, a unified framework comprising two core components: the MFM module, which projects overflowing frames into representative frequency coefficients complemented by salient residuals to reconstruct the global "gist" of historical context; and the STM module, which discretizes continuous streams into episodic clusters and employs an adaptive compression strategy to distill them into high-density space thumbnails.

\textbf{Training-free Versatility with Scalable SOTA Performance.}
As a plug-and-play module, FreshMem delivers immediate training-free gains that surpass many fully fine-tuned methods. Furthermore, its performance can be further amplified through low-cost component-wise fine-tuning to achieve state-of-the-art results, offering a highly efficient and scalable paradigm for streaming video understanding.
\section{Related Work}
\label{sec:related_work}

\subsection{Input-level Modulation for Video Understanding}

In the context of long-form video, static downsampling remains the dominant paradigm. Methods like AKS~\cite{AKS}, and BOLT~\cite{BOLT}
employ heuristic metrics or learnable policies to filter redundant frames based on query relevance, while TimeViper~\cite{TimeViper} adopts linear-complexity architectures to process denser inputs efficiently.
In dynamic streaming scenarios, modulation strategies shift toward managing a sliding computational window.
TimeChat~\cite{ren2024timechat} attends time information by a time-aware frame encoder and a sliding window Q-Former,
while StreamingAssistant~\cite{StreamingAssistant} 
further reduces redundancy by pruning tokens based on spatial adjacency or decoupling positional encodings. 

Current methods, limited by rigid spatial heuristics and uniform windowing, overlook temporal nuances. Consequently, they sacrifice high-frequency details like subtle motion, hindering the reconciliation of local fidelity with the evolving narrative.

\subsection{Memory-augmented Architectures for Video Understanding}

A primary direction involves KV cache compression. LiveVLM~\cite{ning2025livevlm} and InfiniPot-V~\cite{InfiniPot-V} dynamically prune redundant key-value pairs in the streaming cache.
Another direction focuses on global token condensation and external banks. Pioneering works like MovieChat~\cite{song2024moviechat} and MA-LMM~\cite{MA-LMM} condense history into token-based summaries. More recent frameworks enforce structure upon these memories: 
Dispider~\cite{qian2025dispider} features a lightweight proactive streaming video processing module to determine the use of memory, 
and Flash-VStream~\cite{zhang2024flashvstream} employs dual-layer storage. Furthermore, VideoLLaMB~\cite{VideoLLaMB}
introduces recurrent bridges to smooth transitions. 

Current architectures fail to capture continuous temporal dynamics, often yielding fragmented representations caused by heavy spatial or lossy textual dependencies. Our proposed FreshMem overcomes this by balancing the retention of critical long-term information with short-term sensitivity, ensuring robust long-horizon reasoning.
\section{\textit{FreshMem}}
\label{Method}

\subsection{Theory Foundations}
FreshMem is grounded in two fundamental neuroscientific observations regarding temporal perception and episodic replay: the Weber-Fechner Law~\cite{law} of logarithmic perception~\cite{logarithmicperception} and the Sharp-Wave Ripples (SWRs)~\cite{swr} mechanism for memory consolidation~\cite{memory}.

\textbf{Weber-Fechner Law.}
Research in psychophysics suggests that the brain's perception of time is non-uniform~\cite{logarithmicperception}. According to the Weber-Fechner Law, the relationship between the objective magnitude of a physical stimulus and its subjective perception is logarithmic rather than linear~\cite{jean}. 

Formally, let $\Delta t$ be the time interval from the current moment. The perceived information density $I(\Delta t)$ can be modeled as:
\begin{equation}
    I(\Delta t) \propto \frac{1}{k \cdot \ln(\Delta t + \epsilon)}
\end{equation}
where $k$ is a scaling constant and $\epsilon$ prevents singularity.
This biological prior motivates us to project historical context into a domain where information density naturally decays, rather than storing frames with uniform density.

\begin{figure*}[t]
  \centering    \centerline{\includegraphics[width=1.0\textwidth]{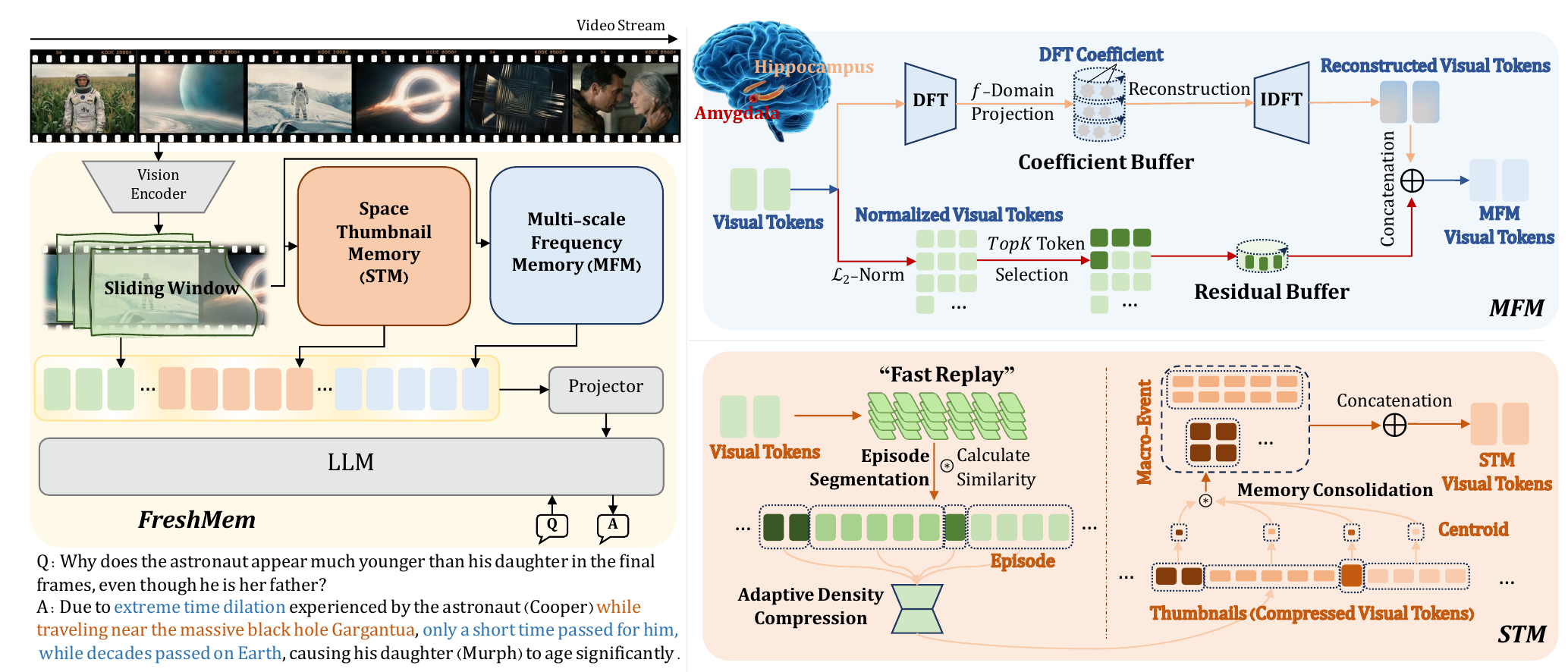}}
    \caption{\textbf{Illustration of FreshMem.} (1) Sliding window for immediate high-fidelity context; (2) MFM, which maintains a global history by frequency coefficients and salient residuals; and (3) STM, which performs online episode segmentation and adaptive compression to keep key spatial thumbnails. The concatenated memory features enable the LLM to answer long-horizon queries with high accuracy.}
    \label{fig2}
\end{figure*}
\textbf{Sharp-Wave Ripples (SWRs).}
While the Weber-Fechner law describes passive decay, the brain also employs active consolidation mechanisms. During slow-wave sleep or rest, the hippocampus generates Sharp-Wave Ripples (SWRs). These high-frequency oscillations support the “fast replay” of sequential experiences, compressing long continuous episodes into brief, high-density neural bursts~\cite{swr}. 

Mathematically, we model this biological process as a Space Compression Operator $\Psi$. Given a continuous stream of visual features $E = \{x_1, x_2, ..., x_T\}$ representing an episode, the SWRs mechanism functions as a non-linear integration that maps the sequence to a high-density “memory trace” $\mathbf{m}$:
\begin{equation}
    \mathbf{m} = \Psi(E) \approx \frac{1}{|E|} \sum_{i=1}^{T} \mathbb{I}(x_i) \cdot x_i
\end{equation}
where $\mathbb{I}(\cdot)$ denotes a saliency filter. This implies that extensive space redundancy is removed, retaining only the topological “centroid” of the event.
This inspires our strategy of dynamically monitoring semantic boundaries to fuse redundant frames into “episodic clusters”, preserving topology within a limited budget.

\subsection{Frequency-Space Hybrid Memory Network} 
Guided by these principles, we propose the FreshMem framework. As illustrated in the left of~\cref{fig2}, video frames are processed by a vision encoder and buffered in a sliding window $\mathcal{M}_{short}$ for immediate context. To retain long-term information, overflowed frames are parallelly assimilated by Multi-scale Frequency Memory (MFM) for global “gist” reconstruction and Space Thumbnail Memory (STM) for episodic clustering. Finally, tokens from these memories are aggregated via a projector and fed into the LLM, enabling robust long-horizon streaming reasoning.
\subsubsection{Multi-scale Frequency Memory (MFM)}
To balance immediate precision with long-term retention, MFM operates in conjunction with sliding window that stores raw features of the most recent frames. As frames overflow, they are not discarded but projected into the frequency domain.
\paragraph{Incremental Frequency Update.}
Standard DFT~\cite{li2026statisticalcharacteristicguideddenoisingrapid,fontana2026faarefficientfrequencyawaremultitask,gu2026fourieranglealignmentoriented,cai2026pfgnetfullyconvolutionalfrequencyguided} requires access to the entire history, which is infeasible for streaming. We implement a recurrent update mechanism. Let $C_t \in \mathbb{C}^{K \times D}$ denote the frequency coefficients at step $t$, where $K$ is the number of frequency bands. For an incoming overflowed frame $x_t$, we update the coefficients incrementally:
\begin{equation}
    C_{t}[k] = \gamma \cdot C_{t-1}[k] + x_t \cdot e^{-j \omega_k t}
\end{equation}
where $\omega_k$ is the angular frequency derived from the Weber-Fechner scale, and $\gamma \in (0, 1)$ is a decay factor preventing saturation. This ensures $O(1)$ update complexity. The coefficients are stored in a circular buffer structure; when the buffer is full, the pointer wraps around, overwriting the oldest frequency interactions, naturally mimicking the fading of very distant memories~\cite{murre2015replication}.

\paragraph{Amygdala-Inspired Residuals.}
Biological memory is not purely semantic; the \textit{Amygdala} ensures that emotionally salient or shocking events are retained with high fidelity~\cite{ledoux2007amygdala}, regardless of their frequency characteristics. Inspired by this, and leveraging observations in~\cite{tenney2019bert} that feature magnitude correlates with information importance, we introduce a \textit{salient residual mechanism}.

During the update, we calculate the $L_2$-norm of the input feature $\|x_t\|_2$ as a proxy for information entropy, which is widely used and proven to be effective in LLM~\cite{xiao2024efficientstreaminglanguagemodels,devoto-etal-2024-simple} and MLLM~\cite{dhouib2025pactpruningclusteringbasedtoken,choi2025representationshiftunifyingtoken,li2025unseenbiasnormdiscrepancy} areas, denoted as $\mathcal{B}_{residual}$. We retain the top-$k$ elements with the highest norm as residual tokens $\mathbf{r}_t$:
\begin{equation}
    \mathbf{r}_t = \begin{cases} 
        x_t, & \text{if } \|x_t\|_2 \in \text{TopK}(\mathcal{B}_{residual}) \\
        \mathbf{0}, & \text{otherwise}
    \end{cases}
\end{equation}
This dual-pathway storage mimics the Hippocampus' global low-frequency context and Amygdala's sparse high-frequency saliency collaboration~\cite{richter2000amygdala}.
\paragraph{Holographic Reconstruction.}
During retrieval at step $\tau$, we reconstruct the history $\hat{H}$ by performing the Inverse Discrete Fourier Transform (IDFT) on the frequency coefficients and fusing them with the retrieved residuals:
\begin{equation}
    \hat{H}_{\tau} = \text{Real}\left( \sum_{k=1}^{K} C_t[k] \cdot e^{j \omega_k \tau} \right) \oplus \mathbf{r}_{\tau}
\end{equation}
This reconstruction allows the model to “hallucinate” the gist of the entire history while preserving sharp details for key frames.
\subsubsection{Space Thumbnail Memory (STM)}
While MFM handles global context, STM is designed to discretize continuous streams into structured “episodic clusters”, capturing the topological evolution of events.
\paragraph{Episode Segmentation.}
Similar to MFM, frames first pass through sliding window. Upon eviction, STM monitors the semantic continuity. We define an episode boundary based on the cosine similarity between the current frame $x_t$ and the preceding context $x_{t-1}$:
\begin{equation}
    \delta(t) = \frac{x_t \cdot x_{t-1}}{|x_t| |x_{t-1}|} < \theta_{event}
\end{equation}
If the similarity score $\delta(t)$ drops below the threshold $\theta_{event}$, a new episodic cluster $E_{new}$ is initialized; otherwise, $x_t$ is appended to the current active episode.

\paragraph{Adaptive Density Compression.}
Within each episode, we apply an adaptive compression strategy to maintain a constant “information density”. The sampling rate $\rho$ is dynamically adjusted based on the episode duration $N$. For an episode $E_i$ with $N$ frames, the stored representation $\mathbf{Z}_i$ is obtained by adaptive downsampling:
\begin{equation}
    \mathbf{Z}_i = \text{Sample}(E_i, \rho(N))
\end{equation}
where the sampling rate $\rho(N) \propto \text{clamp}(N, \rho_{min}, \rho_{max})$ adapts to the episode length.
This ensures that short, fast-paced actions retain higher frame rates, while long, redundant stationary scenes are heavily compressed, optimizing the storage budget.
\paragraph{Centroid-based Memory Consolidation.}
To strictly enforce the memory budget, we propose a topology-aware merge strategy. Instead of simply discarding the oldest event (FIFO), we calculate the \textit{semantic centroid}~\cite{shang2026llavaprumergeadaptivetokenreduction} $\mu_i$ for each stored episode $E_i$. When the memory is full, we compute the pairwise similarity $S_{merge}$ between adjacent event centroids:
\begin{equation}
    S_{merge}(i, i+1) = \frac{\mu_i \cdot \mu_{i+1}}{|\mu_i| |\mu_{i+1}|}
\end{equation}
If $S_{merge}$ exceeds the fusion threshold $\theta_{merge}$, the two episodes are merged into a single “Macro-Event”, summing their frame counts and re-calculating the unified centroid. This mechanism effectively fuses redundant micro-events into a coherent narrative, preserving the deep semantic topology of the video.
\subsubsection{Global Memory Integration}
Finally, the comprehensive memory representation $\mathcal{M}_{total}$ for the FreshMem framework is constructed by concatenating the outputs of all three modules:
\begin{equation}
    \mathcal{M}_{total} = [\mathcal{M}_{MFM} \parallel \mathcal{M}_{STM} \parallel \mathcal{M}_{short}]
\end{equation}
This hybrid token sequence provides the downstream predictor with global historical context, structured episodic summaries and immediate details.
\begin{table*}[t]
    \centering
    \begin{minipage}[t]{0.54\linewidth}
        \centering
        \caption{
            \textbf{Evaluation results on OV-Bench.}
            FP, THV, PM, SP, STP, TP represent Future Prediction, Temporal Hallucination Verification, Past Memory, Spatial, Spatio-Temporal, Temporal Perception respectively.
        }
        \label{tab:ovbench}
        \resizebox{\linewidth}{!}{
            \begin{tabular}{l ccccccc}
                \toprule
                Method & FP & THV & PM & SP & STP & TP & \textbf{Avg.}\\
                \midrule
                \rowcolor{tablehighlight}
                \multicolumn{8}{l}{\textit{Proprietary MLLMs}} \\ \midrule
                Gemini-1.5-Flash~\cite{geminiteam2024gemini15unlockingmultimodal} & 48.97 & 52.63 & 49.03 & 34.90 & 68.50 & 50.83 & 50.7 \\
                \midrule 
                \rowcolor{tablehighlight}
                \multicolumn{8}{l}{\textit{Open-source Offline Video MLLMs}} \\ \midrule
                
                InternVL2-7B~\cite{chen2024internvl2} & 46.80 & 56.00 & 54.73 & 34.70 & 62.20 & 37.57 & 48.7 \\
                InternVL2-4B~\cite{chen2024internvl2} & 43.03 & 56.20 & 46.13 & 25.50 & 57.65 & 34.50 & 44.1 \\
                LLaMA-VID-7B~\cite{li2024llamavid} & 38.03 & 52.77 & 43.17 & 21.55 & 50.25 & 41.70 & 41.9 \\
                LLaVA-Onevision-7B~\cite{li2024llavaonevision} & 55.53 & 51.73 & 49.37 & 28.65 & 68.10 & 43.00 & 49.5 \\
                LongVA-7B~\cite{zhang2024long} & 50.03 & 47.20 & 49.53 & 17.50 & 56.05 & 36.90 & 43.6 \\
                MiniCPM-V2.6-7B~\cite{yao2024minicpm} & 28.07 & 55.03 & 34.70 & 19.20 & 67.30 & 32.90 & 39.1 \\
                LITA-7B~\cite{huang2024lita} & 21.20 & 38.20 & 12.27 & 10.75 & 29.55 & 10.43 & 20.4 \\
                TimeChat-7B~\cite{ren2024timechat} & 13.90 & 16.00 & 11.20 & 13.50 & 12.10 & 10.07 & 12.8 \\
                VTimeLLM-7B~\cite{huang2024vtimellm} & 25.20 & 53.93 & 32.40 & 23.15 & 23.85 & 33.50 & 33.1 \\

                \midrule
                \rowcolor{tablehighlight}
                \multicolumn{8}{l}{\textit{Open-source Online Video MLLMs}} \\ \midrule
                VideoLLM-Online-7B~\cite{chen2024videollmonline} & 7.57 & 14.57 & 9.67 & 13.60 & 10.40 & 3.63 & 9.6 \\
                MovieChat-7B~\cite{song2024moviechat}  & 24.73 & 47.53 & 26.87 & 33.25 & 26.65 & 25.57 & 30.9 \\
                Flash-Vstream-7B~\cite{zhang2024flashvstream} & 29.47 & 47.33 & 28.27 & 26.75 & 24.20 & 27.40 & 31.2 \\
                
                \midrule
                
                InternVL2-2B~\cite{chen2024internvl2}     & 44.36 & 49.66 & 50.24 & 23.64 & 52.72 & 26.44 & 42.06 \\
                \rowcolor{mylightblue}
                \quad\textbf{+\ourmethod}   & 46.62 & 54.13 & 59.69 & 30.64 & 57.79 & 29.76 & \bf{47.44}(\textcolor{red}{$\uparrow$ 5.38\%}) \\
                Qwen2VL-7B~\cite{wang2024qwen2vl}    & 45.39 & 50.47 & 57.80 & 22.71 & 69.02 & 30.66 & 46.30 \\
                \rowcolor{mylightblue}
                \quad\textbf{+\ourmethod}     & 50.38 & 59.73 & 60.87 & 25.51 & 72.64 & 32.39 & \bf{50.82}(\textcolor{red}{$\uparrow$ 4.52\%}) \\
                \bottomrule
            \end{tabular}
        }
    \end{minipage}
    \hfill 
    \begin{minipage}[t]{0.45\linewidth}
        \centering
        \caption{
            \textbf{Evaluation on OVO-Bench.} RTVP, BT, FAR denote Real-Time Visual Perception, Backward Tracing, Forward Active Responding.
        }
        \label{tab:ovobench-small}
        \resizebox{\linewidth}{!}{
            \begin{tabular}{l cccc}
                \toprule
                Method & RTVP & BT & FAR & \textbf{Avg.}\\
                \midrule
                \rowcolor{tablehighlight}
                \multicolumn{5}{l}{\textit{Proprietary MLLMs}} \\ \midrule
                Gemini 1.5 Pro~\cite{geminiteam2024gemini15unlockingmultimodal} & 69.32 & 62.54 & 57.15 & 63.00 \\
                GPT-4o~\cite{hurst2024gpt} & 64.46 & 60.75 & 53.40 & 59.54 \\
                \midrule
                \rowcolor{tablehighlight}
                \multicolumn{5}{l}{\textit{Open-source Offline Video MLLMs}} \\ \midrule
                LLaVA-Video~\cite{zhang2024llavavideo} & 63.34 & 41.68 & 54.17 & 53.06 \\
                LLaVA-OneVision~\cite{li2024llavaonevision} & 62.79 & 44.99 & 50.85 & 52.88 \\
                InternVL2-8B~\cite{chen2024internvl2} & 60.73 & 44.00 & 45.42 & 50.05 \\
                LongVU-7B~\cite{shen2024longvu} & 57.40 & 39.49 & 48.54 & 48.48 \\

                \midrule
                \rowcolor{tablehighlight}
                \multicolumn{5}{l}{\textit{Open-source Online Video MLLMs}} \\ \midrule
                VideoLLM-online~\cite{chen2024videollmonline} & 20.79 & 17.73 & - & 12.84\\
                Flash-VStream~\cite{zhang2024flashvstream} & 29.86 & 25.35 & 44.23 & 33.15 \\
                Dispider-7B~\cite{qian2025dispider} & 54.55 & 36.06 & 34.72 & 41.78 \\
                \midrule
                InternVL2-2B~\cite{chen2024internvl2}    & 44.92 & 34.55 & 41.93 & 41.22 \\
                \rowcolor{mylightblue}
                \quad\textbf{+\ourmethod}   & 47.79 & 40.10 & 42.63 & \bf{43.53}(\textcolor{red}{$\uparrow$ 2.31\%}) \\
                \midrule
                Qwen2VL-7B~\cite{wang2024qwen2vl}    & 60.65 & 49.13 & 48.86 & 52.19 \\
                \rowcolor{mylightblue}
                \quad\textbf{+\ourmethod}   & 66.67 & 54.04 & 48.25 & \bf{54.53}(\textcolor{red}{$\uparrow$ 2.34\%}) \\
                \bottomrule
            \end{tabular}
        }
    \end{minipage}
\end{table*}
\section{Experiments}
\label{Experiments}

\begin{table*}[t]
\centering
\caption{
\textbf{Evaluation results on StreamingBench.}
}
\label{tab:streamingbench}
\footnotesize
\resizebox{1.0\linewidth}{!}{
\begin{tabular}{l ccccccccccc}
    \toprule
    Method & OP & CR & CS & ATP & EU & TR & PR & SU & ACP & CT & ALL\\
    \midrule
    \rowcolor{tablehighlight}
    \multicolumn{12}{l}{\textit{Proprietary MLLMs}} \\
    \midrule
    Gemini 1.5 Pro~\cite{geminiteam2024gemini15unlockingmultimodal} & 79.02 & 80.47 & 83.54 & 79.67 & 80.00 & 84.74 & 77.78 & 64.23 & 71.95 & 48.70 & 75.69 \\
    GPT-4o~\cite{hurst2024gpt} & 77.11 & 80.47 & 83.91 & 76.47 & 70.19 & 83.80 & 66.67 & 62.19 & 69.12 & 49.22 & 73.28 \\
    Claude 3.5 Sonnet~\cite{anthropic2024claude} & 73.33 & 80.47 & 84.09 & 82.02 & 75.39 & 79.53 & 61.11 & 61.79 & 69.32 & 43.09 & 72.44 \\

    \midrule    
    \rowcolor{tablehighlight}
    \multicolumn{12}{l}{\textit{Open-source Offline Video MLLMs}} \\
    \midrule
    Video-LLaMA2-7B~\cite{cheng2024videollama} & 55.86 & 55.47 & 57.41 & 58.17 & 52.80 & 43.61 & 39.81 & 42.68 & 45.61 & 35.23 & 49.52 \\
    LongVA-7B~\cite{zhang2024long} & 70.03 & 63.28 & 61.20 & 70.92 & 62.73 & 59.50 & 61.11 & 53.66 & 54.67 & 34.72 & 59.96 \\
    InternVL2-8B~\cite{chen2024internvl2} & 68.12 & 60.94 & 69.40 & 77.12 & 67.70 & 62.93 & 59.26 & 53.25 & 54.96 & 56.48 & 63.72 \\
    Kangaroo-7B~\cite{liu2024kangaroo} & 71.12 & 84.38 & 70.66 & 73.20 & 67.08 & 61.68 & 56.48 & 55.69 & 62.04 & 38.86 & 64.60 \\
    LLaVA-NeXT-Video-32B~\cite{zhang2024llavavideo} & 78.20 & 70.31 & 73.82 & 76.80 & 63.35 & 69.78 & 57.41 & 56.10 & 64.31 & 38.86 & 66.96 \\
    MiniCPM-V2.6-8B~\cite{yao2024minicpm} & 71.93 & 71.09 & 77.92 & 75.82 & 64.60 & 65.73 & 70.37 & 56.10 & 62.32 & 53.37 & 67.44 \\
    LLaVA-OneVision-7B~\cite{li2024llavaonevision} & 80.38 & 74.22 & 76.03 & 80.72 & 72.67 & 71.65 & 67.59 & 65.45 & 65.72 & 45.08 & 71.12 \\
    Qwen2.5-VL-7B~\cite{bai2025qwen2} & 78.32 & 80.47 & 78.86 & 80.45 & 76.73 & 78.50 & 79.63 & 63.41 & 66.19 & 53.19 & 73.68 \\

    \midrule
    \rowcolor{tablehighlight}
    \multicolumn{12}{l}{\textit{Open-source Online Video MLLMs}} \\
    \midrule
    Flash-VStream-7B~\cite{zhang2024flashvstream}  &25.89 & 43.57 & 24.91 & 23.87 & 27.33 & 13.08 & 18.52 & 25.20 & 23.87 & 48.70 & 23.23 \\
    VideoLLM-online-8B~\cite{chen2024videollmonline} & 39.07 & 40.06 & 34.49 & 31.05 & 45.96 & 32.40 & 31.48 & 34.16 & 42.49 & 27.89 & 35.99 \\
    Dispider-7B~\cite{qian2025dispider} & 74.92 & 75.53 & 74.10 & 73.08 & 74.44 & 59.92 & 76.14 & 62.91 & 62.16 & 45.80 & 67.63 \\
    ReKV-7B~\cite{di2025rekv} & 74.39 & 78.91 & 78.55 & 77.12 & 68.32 & 67.91 & 67.59 & 62.60 & 64.31 & 44.56 & 69.08\\
    LiveVLM-7B~\cite{ning2025livevlm} & 81.47 & 78.13 & 83.28 & 79.08 & 69.57 & 74.14 & 75.00 & 69.11 & 67.71 & 40.41 & 72.92 \\
    StreamForest~\cite{zeng2025streamforest} & 83.11& 82.81& 82.65& 84.26& 77.50 &78.19 &76.85 &69.11& 75.64& 54.40& 77.26\\

    \midrule
    
    InternVL2-2B~\cite{chen2024internvl2}      & 58.58 & 50.78 & 59.62 & 66.89 & 57.50 & 49.22 & 52.78 & 48.37 & 47.31 & 18.13 & 52.08 \\
    \rowcolor{mylightblue}
    \quad\textbf{+\ourmethod}   & 64.85 & 57.81 & 66.56 & 72.46 & 63.75 & 58.26 & 60.19 & 52.44 & 53.82 & 19.17 & 58.21(\textcolor{red}{$\uparrow$ 6.13\%}) \\
    \rowcolor{mylightblue}
    \quad\textbf{+\ourmethod(ViT fine-tuned)}            & 62.47 & 62.13 & 65.82 & 69.29 & 67.54 & 63.88 & 66.03 & 59.71 & 60.35 & 24.56 & 60.44(\textcolor{red}{$\uparrow$ 8.36\%}) \\
    \rowcolor{mylightblue}
    \quad\textbf{+\ourmethod(Projector fine-tuned)}            & 63.74 & 63.29 & 65.81 & 70.33 & 68.57 & 64.98 & 67.12 & 59.46 & 60.03 & 25.67 & 60.82(\textcolor{red}{$\uparrow$ 8.74\%}) \\
    \rowcolor{mylightblue}
    \quad\textbf{+\ourmethod(LLM fine-tuned)}            & 62.39 & 66.12 & 65.87 & 70.54 & 71.63 & 67.98 & 69.45 & 61.26 & 61.08 & 27.31 & \bf{62.69}(\textcolor{red}{$\uparrow$ 10.61\%}) \\
    Qwen2VL-7B~\cite{wang2024qwen2vl}    & 77.38 & 76.56 & 73.19 & 75.08 & 75.00 & 67.91 & 73.15 & 65.04 & 66.57 & 35.75 & 69.00 \\
    \rowcolor{mylightblue}
    \quad\textbf{+\ourmethod}            & 84.47 & 83.59 & 77.60 & 83.28 & 78.12 & 80.37 & 70.37 & 74.39 & 66.86 & 30.05 & 74.20(\textcolor{red}{$\uparrow$ 5.20\%}) \\
    \rowcolor{mylightblue}
    \quad\textbf{+\ourmethod(ViT fine-tuned)}            & 82.24 & 85.29 & 78.10 & 82.25 & 81.68 & 78.27 & 75.85 & 77.72 & 73.64 & 36.54 & 76.35(\textcolor{red}{$\uparrow$ 7.35\%}) \\
    \rowcolor{mylightblue}
    \quad\textbf{+\ourmethod(Projector fine-tuned)}            & 84.93 & 82.28 & 80.63 & 84.26 & 82.74 & 78.22 & 77.18 & 79.63 & 74.75 & 36.47 & 77.73(\textcolor{red}{$\uparrow$ 8.73\%}) \\
    \rowcolor{mylightblue}
    \quad\textbf{+\ourmethod(LLM fine-tuned)}            & 82.65 & 90.32 & 90.18 & 83.79 & 85.01 & 85.64 & 84.36 & 76.62 & 83.14 & 58.48 & \bf{82.31}(\textcolor{red}{$\uparrow$ 13.31\%}) \\
    \bottomrule
\end{tabular}
}
\end{table*}

We employ Qwen2-VL-7B~\cite{wang2024qwen2vl} and InternVL2-2B~\cite{chen2024internvl2} as baselines and test the performance on both online benchmarks (see ~\cref{Online results}) and offline benchmarks (see ~\cref{Offline results}).
FreshMem is training-free and it can be directly applied on top of baseline video MLLMs during inference. All experiments are conducted on RTX4090 and H200 GPUs. Detailed settings for all parameters are provided in~\cref{tab:settings}.

\subsection{Online Benchmark Results}
\label{Online results}
We evaluate the performance of our model on three benchmarks for online video question answering: OV-Bench~\cite{huang2025ovbench}, OVO-Bench~\cite{niu2025ovobench}, and StreamingBench~\cite{lin2024streamingbench}. 
These benchmarks are formulated as multiple-choice questions. For each question, the answer must be generated solely based on the video content before the corresponding timestamp, without access to the entire video.

Experimental results of these three benchmarks are shown in~\cref{tab:ovbench,tab:ovobench-small,tab:streamingbench}. FreshMem achieves stable and significant performance improvement over different baselines and benchmarks. Built upon Qwen2-VL-7B, it obtains an accuracy of 50.82\% on OV-Bench, 54.53\% on OVO-Bench, and 74.2\% on StreamingBench, yielding absolute improvements of 4.52\%, 2.34\%, and 5.20\% over the baseline, respectively. 
This improvement arises from FreshMem’s brain-inspired memory strategy, which retains recent details with a frequency domain projection while compressing long-term context into semantic gist and episodic thumbnails. Notably, our method outperforms several supervised approaches, despite not requiring extensive training. By aligning memory retention with video temporal structure, our method preserves both short-term precision and long-term coherence without additional training. More Experiments on InternVL3.5-7B and DeepSeek-VL2-Tiny are in~\cref{tab:moremodel}.

To explore trainable extensions of FreshMem, following the end-to-end training settings established by StreamForest~\cite{zeng2025streamforest}, we independently fine-tuned the ViT, the Projector, and the LLM components of our architecture. Results in~\cref{tab:streamingbench} show FreshMem achieves a dominant 82.31\%, vastly outperforming both our training-free baseline and the fully trained StreamForest model and reaching state-of-the-art performance.

\subsection{Offline Benchmark Results}
\label{Offline results}

We further evaluate our method on one long video understanding benchmark MLVU~\cite{zhou2025mlvu} and one short video benchmark MVBench~\cite{li2024mvbench} and report the results in~\cref{tab:mlvu+mvbench}. In these offline
settings, the MLLM is given access to the complete video as input. Our method demonstrates improvement upon the Qwen2VL-7B baseline by 2.4\% on MLVU and by 4.7\% on MVBench. These results highlight that our method not only excels on online benchmarks but also generalizes effectively to offline benchmarks.

\begin{table}[htbp]
\centering
\begin{minipage}{0.53\linewidth}
\centering
\caption{\textbf{Evaluation results on MLVU and MVBench.}}
\label{tab:mlvu+mvbench}
\footnotesize
\resizebox{\linewidth}{!}{
\begin{tabular}{l cc}
    \toprule
    Method & MLVU & MVBench \\
    \midrule    
    LongVA-7B~\cite{zhang2024long} & 56.3 & - \\
    LLaVA-OneVision-7B~\cite{li2024llavaonevision} & 64.7 & 56.7 \\
    LLaVA-Video-7B~\cite{zhang2024llavavideo} & 70.8 & 58.6 \\
    MovieChat-7B~\cite{song2024moviechat}  & - & 55.1 \\
    \midrule
    Qwen2VL-7B~\cite{wang2024qwen2vl}    & 54.9 & 59.1 \\
    \rowcolor{mylightblue}
    \quad\textbf{+\ourmethod}   & \bf{57.3}(\textcolor{red}{$\uparrow$ 2.4\%}) & \bf{63.8}(\textcolor{red}{$\uparrow$ 4.7\%}) \\
    \bottomrule
\end{tabular}
}
\end{minipage}
\hfill
\begin{minipage}{0.45\linewidth}
\centering
\caption{\textbf{Ablation study on OVO-Bench results.} SW represents Sliding Window.}
\label{tab:ablation1}
\footnotesize
\resizebox{\linewidth}{!}{
\begin{tabular}{lcccc}
    \toprule
    \textbf{Model} & \textbf{SW} & \textbf{STM} & \textbf{MFM} & \textbf{Avg.} \\
    \midrule
    Qwen2-VL-7B & $\times$ & $\times$ & $\times$ & 52.19 \\
    \midrule
    \rowcolor{tablehighlight}
    + SW   & \checkmark & $\times$ & $\times$ & 52.52 \\
    \rowcolor{tablehighlight}
    + STM              & $\times$ & \checkmark & $\times$ & 53.64 \\
    \rowcolor{tablehighlight}
    + MFM              & $\times$ & $\times$ & \checkmark & 53.54 \\
    \rowcolor{tablehighlight}
    + SW + STM         & \checkmark & \checkmark & $\times$ & 53.92 \\
    \rowcolor{tablehighlight}
    + SW + MFM         & \checkmark & $\times$ & \checkmark & 54.36 \\
    \rowcolor{tablehighlight}
    + STM + MFM        & $\times$ & \checkmark & \checkmark & 54.05 \\
    \midrule
    \rowcolor{mylightblue}
    \textbf{Ours}      & \checkmark & \checkmark & \checkmark & \textbf{54.53} \\
    \bottomrule
\end{tabular}
}
\end{minipage}
\end{table}

\begin{figure}[h]
  \centering    \centerline{\includegraphics[width=0.8\linewidth]{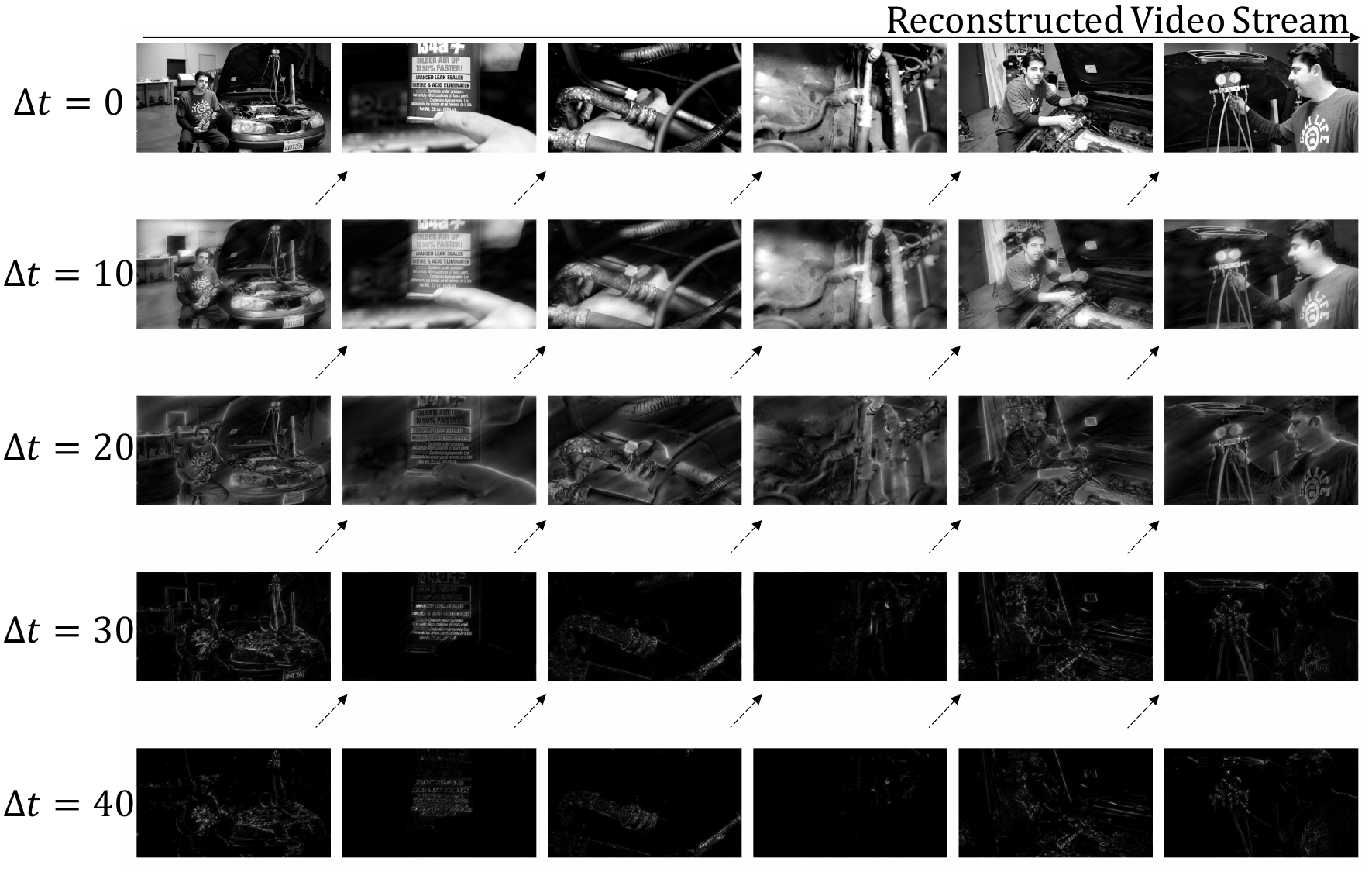}}
    \caption{\textbf{Reconstructed video stream from frequency coefficients.} The visualization demonstrates the information decay inherent in our MFM module.}
    \label{vis1}
\end{figure}

\subsection{Visualization and Case Study}
To enhance the interpretability of FreshMem, we visualized the reconstructed visual tokens during the streaming video process by decoding them back into complete video frames, as illustrated in~\cref{vis1}. As observed in the figure, frames further from the current timestamp ($\Delta t$) appear increasingly blurred and tend towards a binary-like representation. This demonstrates that following DFT and IDFT processing, the MFM effectively preserves the “gist” information—maintaining a rough outline of past events that gradually decays over time. This phenomenon empirically validates the rationality of the logarithmic perception mechanism inherent in our MFM module.

\begin{figure}[t]
  \centering    \centerline{\includegraphics[width=0.8\linewidth]{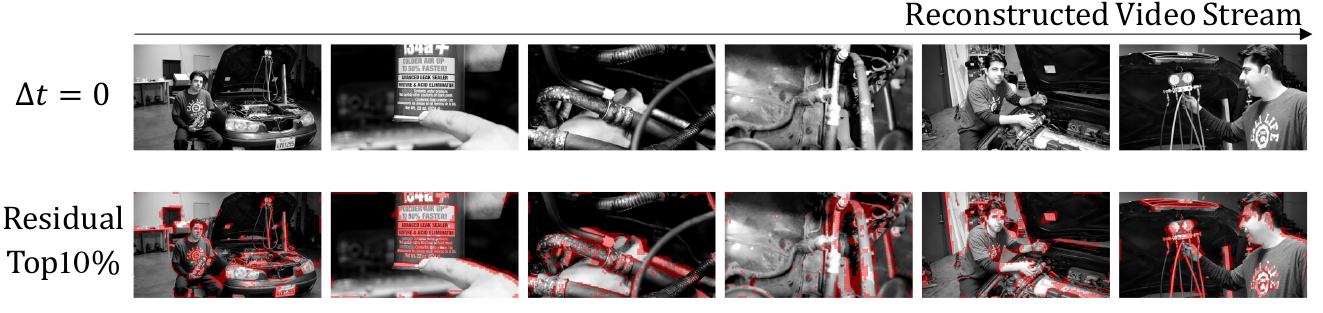}}
    \caption{\textbf{Impact of Salient Residual Tokens.} We visualize the reconstruction of the top-10\% residual tokens selected by the MFM.}
    \label{vis2}
\end{figure}

Furthermore, we visualized the reconstruction of the top-$k$ residual tokens selected within the MFM, as shown in~\cref{vis2}. The visualization reveals that the selected top-$k$ tokens are predominantly distributed along the edges of people and objects, clustering in semantically dense regions within each frame. This empirically validates that these residual tokens serve as a vital complement to the reconstructed visual tokens. Specifically, while the reconstructed visual tokens for timestamps distant from the present tend to fade into a blurred background, the residual tokens effectively delineate sharp contours and structural details on top of them. Together, these two components work synergistically to achieve robust memory and recall of past events.

\begin{figure*}[t]
  \centering    \centerline{\includegraphics[width=1.0\textwidth]{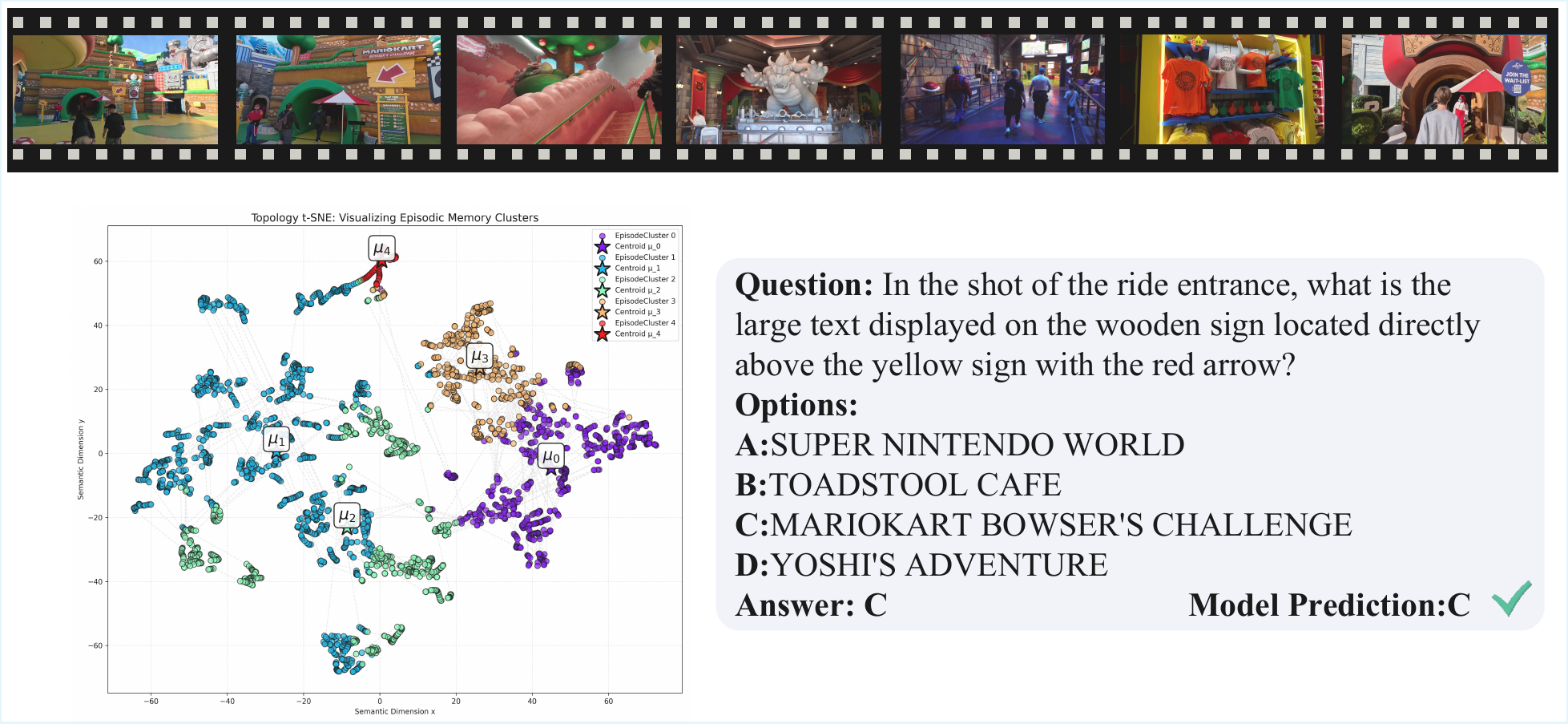}}
    \caption{\textbf{Interpretability analysis of the STM module.} The t-SNE visualization illustrates the effectiveness of our Memory Consolidation mechanism, showing how continuous video streams are discretized into coherent episodic clusters.}
    \label{vis3}
\end{figure*}

Finally, we conduct an interpretability experiment and a case study specifically for the STM module, as illustrated in~\cref{vis3}. We visualized the “episode clusters” using t-SNE and annotated the centroid ($\mu$) of each cluster. In conjunction with the provided case study, we can observe that although the streaming video contains frames that are semantically similar but temporally discontinuous, our proposed episode segmentation and cluster fusion mechanisms enable the model to effectively capture these underlying temporal correlations. This empirically validates that our STM module mimics the biological process of memory consolidation, effectively reducing the semantic fragmentation of visual information.

\subsection{Ablation Studies}
\label{ablation}

To evaluate the effectiveness of each component in our proposed method, we conduct a series of ablation experiments on the OVO-Bench. As shown in~\cref{tab:ablation1}, the baseline Qwen2-VL-7B model achieves an average score of 52.19\%. By individually integrating the Sliding Window (SW), STM, and MFM, we observe consistent performance gains of 0.33\%, 1.45\%, and 1.35\% respectively, demonstrating the necessity of each module. Combining these components further boosts the performance, and our full configuration (SW + STM + MFM) achieves the best result of 54.53\%, which is a 2.34\% absolute improvement over the baseline.

\section{Conclusion}
\label{Conclusion}

In this paper, we proposed FreshMem, a brain-inspired framework that reconciles short-term fidelity with long-term coherence via frequency-domain projection and episodic clustering. Extensive experiments demonstrate that FreshMem achieves distinguished performance on multiple streaming benchmarks. Remarkably, despite being a training-free solution, it surpasses several fully fine-tuned methods, highlighting the superior efficacy of our structured memory design. Following low-cost component-wise fine-tuning, FreshMem provides crucial inspiration for designing more reliable parameterized memory structures and cost-effective training recipes. We firmly believe that FreshMem serves as a vital bridge, transitioning multimodal models from heuristic, non-parametric memory designs toward highly efficient, trainable parametric memory paradigms.

\begin{ack}
This work is supported by National Natural Science Foundation of China (No. 62571138), National Key R\&D Program of China (No. 2026YFE0101200), and National Natural Science Foundation of China (No. 62671175). The computations in this research were performed using the CFFF platform of Fudan University.

\end{ack}

{
    \small
    \bibliographystyle{abbrv}
    \bibliography{main/main}
}


\appendix

\appendix
\onecolumn
\section{Inference Efficiency and Memory Analysis}
\label{eff}
To quantitatively assess the computational and memory overhead introduced by our proposed FreshMem, we conducted rigorous inference time and memory usage measurements on the Qwen2-VL-7B model using the AVA subset from OV-Bench —a representative benchmark for streaming video understanding. Our evaluation focuses on absolute inference latency, peak GPU memory usage, and the corresponding performance gains. The results are summarized in~\cref{tab:efficiency_analysis}.

\begin{table}[h]
    \centering
    \caption{\textbf{Inference Efficiency Analysis on OV-Bench (AVA Subset).} Comparison of inference latency, peak GPU memory usage, and accuracy between the baseline and our proposed method. FreshMem achieves a significant performance boost (+1.79\%) with a moderate and acceptable increase in computational overhead, ensuring deployment feasibility on consumer-grade GPUs (e.g., RTX 4090).}
    \label{tab:efficiency_analysis}
    \setlength{\tabcolsep}{10pt} 
    \begin{tabular}{l c c c}
        \toprule
        \textbf{Method} & \textbf{Latency} & \textbf{GPU Mem.} & \textbf{Accuracy (\%)} \\
        \midrule
        Qwen2-VL-7B & 1h 21min & 17GB & 45.63 \\
        \textbf{+ FreshMem (Ours)} & 1h 35min & 20GB & \textbf{47.42} \\
        \bottomrule
    \end{tabular}
\end{table}

\paragraph{Controlled Computational and Memory Overhead.}
Compared to the vanilla Qwen2-VL-7B baseline, FreshMem introduces a moderate and highly controlled increase in resource consumption. Specifically, the relative time overhead is approximately $\times 1.17$ (increasing from 81 mins to 95 mins), and the GPU memory usage sees a marginal increase of roughly $17.6\%$ (from 17GB to 20GB). Crucially, the peak memory footprint of 20GB remains well within the capacity of standard consumer-grade GPUs (e.g., RTX 3090/4090), ensuring that our method maintains broad hardware accessibility and operational feasibility without requiring specialized high-end infrastructure.
\paragraph{Breakdown of Overhead Sources.}
The primary computational overhead in our method stems from the specific memory operations: the incremental Discrete Fourier Transform (DFT) updates in the MFM module and the centroid-based clustering in the STM module. However, these components are mathematically designed to be lightweight with near-linear complexity regarding the window size. By avoiding the heavy retraining of the entire model, FreshMem minimizes the computational burden typically associated with long-context adaptation.
\paragraph{Efficiency-Performance Trade-off.}
The marginal additional resource consumption is well justified by the substantial performance gains observed. While incurring only a slight overhead in latency and memory, FreshMem achieves a significant performance boost, improving accuracy from $45.63\%$ to $47.42\%$ ($+1.79\%$). This highlights that our approach strikes a highly favorable balance between computational complexity and memory-enhanced modeling capability. For latency-sensitive streaming applications, FreshMem offers a high-yield upgrade path that significantly enhances long-horizon reasoning with negligible impact on system throughput.
\paragraph{Summary.}
FreshMem maintains operational feasibility with predictable computational costs. It integrates seamlessly with the base model, providing a superior performance-efficiency trade-off that makes it suitable for both research exploration and real-world deployment scenarios where resources are constrained but high performance is demanded.

\section{Detailed Hyperparameter Settings}
\begin{figure*}[htbp]
  \centering    \centerline{\includegraphics[width=\textwidth]{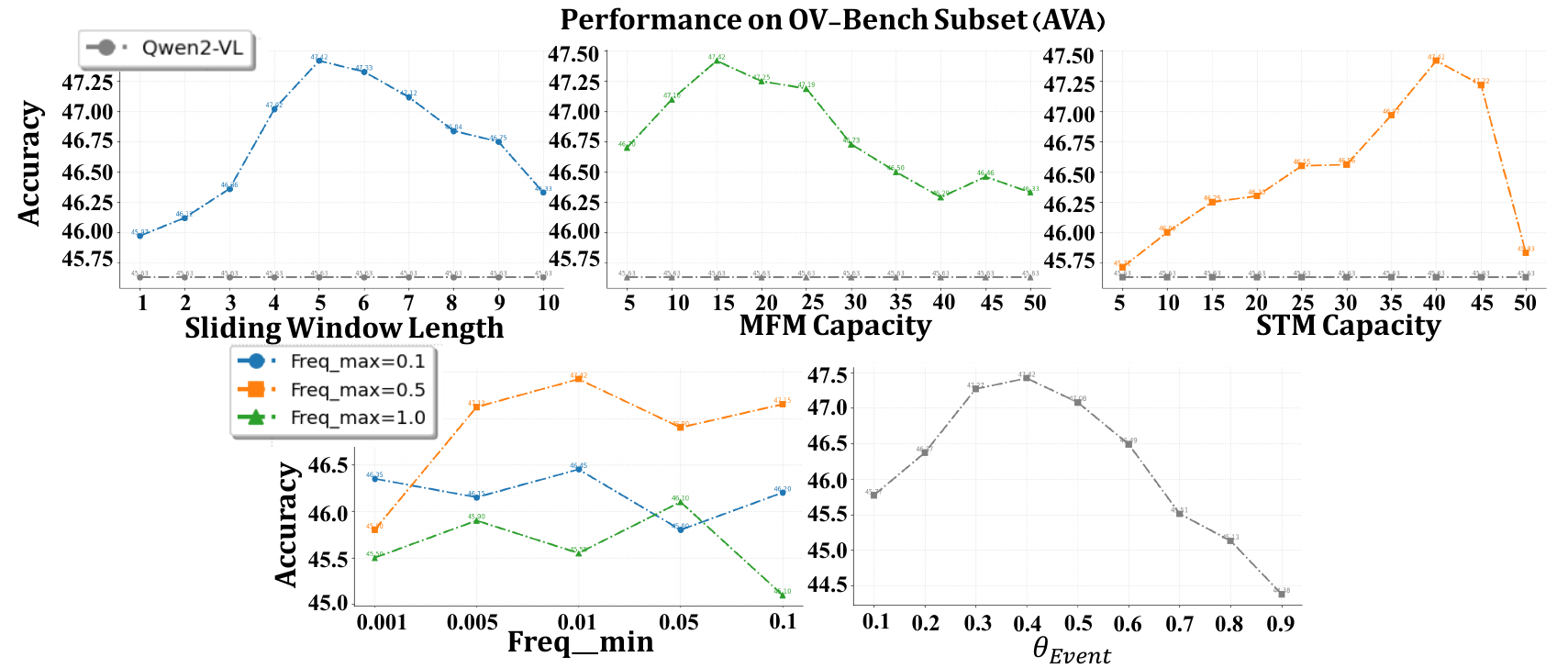}}
    \caption{\textbf{Sensitivity analysis of FreshMem hyperparameters.} Experiments on the OV-Bench subset demonstrate the robustness of our default settings.}
    \label{ab}
\end{figure*}

To verify the impact of key hyperparameters and determine the optimal configuration, we conduct a series of experiments on the OV-Bench Subset (AVA). The results are illustrated in~\cref{ab}, covering sliding window length, memory module capacities (MFM and STM), frequency band, and the episode segmentation threshold $\theta_{event}$.
As shown in the top-left of~\cref{ab}, performance peaks when the sliding window length is set to approximately 5 frames. A shorter window fails to capture sufficient temporal context, while an excessively long window introduces redundancy or noise, leading to performance degradation. 
Then we analyzed the capacity settings for MFM and STM modules in the top-middle and top-right  plots of~\cref{ab}. For MFM, performance reaches optima at capacities of 15. For STM, accuracy improves with capacity, saturating and achieving near-optimal performance around a capacity of 40. Considering the trade-off between inference GPU memory usage, computational latency, and model accuracy, we prioritized efficiency.

The bottom-left plot in~\cref{ab} illustrates the model's 
sensitivity to frequency band parameters ($Freq\_min$ and $Freq\_max$). The orange curve indicates that setting the high-frequency cutoff ($Freq\_max$) to 0.5 yields the best overall performance. Furthermore, we observed that the model extracts the most discriminative temporal features when the high and low frequencies differ by approximately one order of magnitude.

Finally, we investigated the influence of the event update threshold, $\theta_{event}$, within the STM module (~\cref{ab}, bottom-right). This parameter controls the model's sensitivity in distinguishing different episodes. The data shows that the model achieves a stable peak performance when $\theta_{event}$ is set between 0.3 and 0.4. Values outside this range resulted in a notable drop in accuracy.
\begin{table}[H]
\centering
\caption{\textbf{Hyperparameter settings in our experiments.}}
\label{tab:settings}
\begin{tabular}{l|c}
\hline
\textbf{Hyperparameter} & \textbf{Value} \\
\hline
Sliding Window Length & 5 \\
MFM Capacity & 15 \\
STM Capacity & 40 \\
Number of Frequency Bands & 16 \\
Frequency Range & $0.01 \sim 0.5$ \\
Decay Factor ($\gamma$) & 0.9 \\
Residual Tokens Ratio & Top 10\% \\
Event Threshold ($\theta_{\text{event}}$) & 0.4 \\
Merge Threshold ($\theta_{\text{merge}}$) & 0.3 \\
Sampling Rate ($\rho$) & $1/16 \sim 1/4$ \\
\hline
\end{tabular}
\end{table}

\section{Detailed Experimental Performance}
We present the full evaluation results on OVO-Bench and MLVU in~\cref{tab:ovobench-appendix,tab:mlvu-appendix}. ~\Cref{tab:ovobench-appendix} shows the full evaluation results on OVO-Bench with all 12 subtasks. FreshMem demonstrates significant improvements over the Qwen2VL-7B baseline and outperforms many offline and online video MLLMs. Improvement on a specific subtask is up to 11.45\%. ~\Cref{tab:mlvu-appendix} shows the full evaluation results on MLVU with all 7 MCQA subtasks. FreshMem outperforms baseline by 2.4\% overall and up to 8.7\% on a specific subtask.
\begin{table*}[h]
    \centering
    \renewcommand{\arraystretch}{1.1}  
    
        \caption{Full Evaluation results on OVO-Bench.}
        
    \resizebox{1.0\textwidth}{!}{%
        \begin{tabular}{lcccccccccccccccc}
            \toprule
            \multicolumn{1}{c}{} &
              \multicolumn{7}{c}{{Real-Time Visual Perception}} &
              \multicolumn{4}{c}{{Backward Tracing}} &
              \multicolumn{4}{c}{{Forward Active Responding}} &
             \\ \cmidrule(lr){2-8} \cmidrule(lr){9-12} \cmidrule(lr){13-16}
            \multicolumn{1}{c}{\multirow{-2}{*}{{Model}}} &
              OCR &
              ACR &
              ATR &
              STU &
              FPD &
              \multicolumn{1}{c}{OJR} &
              \multicolumn{1}{c}{Avg.} &
              EPM &
              ASI &
              \multicolumn{1}{c}{HLD} &
              \multicolumn{1}{c}{Avg.} &
              REC &
              SSR &
              \multicolumn{1}{c}{CRR} &
              \multicolumn{1}{c}{Avg.} &
              \multirow{-2}{*}{Overall Avg.} \\ \midrule

            \rowcolor{tablehighlight}\multicolumn{17}{l}{\textit{Offline Video MLLMs}} \\ \midrule

            Gemini 1.5 Pro~\cite{geminiteam2024gemini15unlockingmultimodal} & 85.91 & 66.97 & 79.31 & 58.43 & 63.37 & 61.96 & 69.32 & 58.59 & 76.35 & 52.64 & 62.54 & 35.53 & 74.24 & 61.67 & 57.15 & 63.00 \\
            GPT-4o~\cite{hurst2024gpt} & 69.80 & 64.22 & 71.55 & 51.12 & 70.30 & 59.78 & 64.46 & 57.91 & 75.68 & 48.66 & 60.75 & 27.58 & 73.21 & 59.40 & 53.40 & 59.54 \\
            \midrule
            \rowcolor{tablehighlight}\multicolumn{17}{l}{\textit{Offline Video MLLMs}}   \\ \midrule

            LLaVA-Video-7B~\cite{zhang2024llavavideo} & 69.80 & 59.63 & 66.38 & 50.56 & 72.28 & 61.41 &  63.34 & 51.18 & 64.19 & 9.68 & 41.68 & 34.10 & 67.57 & 60.83 & 54.17 & 53.06 \\
            LLaVA-OneVision-7B~\cite{li2024llavaonevision} & 67.11 & 58.72 & 69.83 & 49.44 & 71.29 & 60.33 & 62.79 & 52.53 & 58.78 & 23.66 & 44.99 & 24.79 & 66.93 & 60.83 & \bf{50.85} & 52.88 \\
            InternVL2-8B~\cite{chen2024internvl2} & 68.46 & 58.72 & 68.97 & 44.94 & 67.33 & 55.98 & 60.73 & 43.10 & 61.49 & 27.41 & 44.00 & 25.79 & 57.55 & 52.92 & 45.42 & 50.05 \\
            LongVU-7B~\cite{shen2024longvu} & 55.70 & 49.54 & 59.48 & 48.31 & 68.32 & 63.04 & 57.40 & 43.10 & 66.22 & 9.14 & 39.49 & 16.62 & 69.00 & 60.00 & 48.54 & 48.48 \\

            \midrule
            
            \rowcolor{tablehighlight}\multicolumn{17}{l}{\textit{Online Video MLLMs}} \\ \midrule
            VideoLLM-online-8B~\cite{chen2024videollmonline} & 8.05 & 23.85 & 12.07 & 14.04 & 45.54 & 21.20 & 20.79 & 22.22 & 18.80 & 12.18 & 17.73 & - & - & - & - & 12.84 \\
            
            Flash-VStream-7B~\cite{zhang2024flashvstream} & 25.50 & 32.11 & 29.31 & 33.71 & 29.70 & 28.80 & 29.86 & 36.36 & 33.78 & 5.91 & 25.35 & 5.44 & 67.25 & 60.00 & 44.23 & 33.15 \\

            Dispider-7B~\cite{qian2025dispider} & 57.72 & 49.54 & 62.07 & 44.94 & 61.39 & 51.63 & 54.55 & 48.48 & 55.41 & 4.30 & 36.06 & 18.05 & 37.36 & 48.75 & 34.72 & 41.78 \\
            
            \midrule
            
            Qwen2VL-7B~\cite{wang2024qwen2vl} & 69.13 & 53.21 & 63.79 & 50.56 & 66.34 & 60.87 & 60.65 & 45.79 & 48.65 & 54.84 & 49.13 & 30.09 & 65.66 & 50.83 &  48.86 & 52.19 \\
            \rowcolor{mylightblue}\quad+\bf{\ourmethod} & 77.18 & 60.55 & 70.69 & 56.74 & 63.37 & 70.65 & 66.67 & 57.24 & 56.76 & 46.77 & 54.04 & 23.93 & 72.66 & 55.00 & 48.25 & \bf{54.53}(\textcolor{red}{$\uparrow$ 2.34\%}) \\
            \bottomrule
        \end{tabular}
        }
    \label{tab:ovobench-appendix}
\end{table*}
\begin{table*}[h]
\centering
\caption{
Full Evaluation results on MLVU.
}
\label{tab:mlvu-appendix}
\footnotesize
\resizebox{0.8\linewidth}{!}{
\begin{tabular}{l cccccccc}
    \toprule
    Method & TR & AR & NQA & ER & PQA & AO & AC & AVG\\
    \midrule   
    MovieChat~\cite{song2024moviechat}  & 18.7 & 10.3 & 23.3 & 15.1 & 16.0 & 17.1 & 15.0 & 16.5 \\
    MA-LMM~\cite{MA-LMM} & 44.0 & 23.1 & 13.3 & 30.2 & 14.0 & 18.6 & 13.3 & 22.4 \\
    LLaVa-OneVision-7B~\cite{li2024llavaonevision} & 83.5 & 56.4 & 46.7 & 58.4 & 58.0 & 35.7 & 23.3 & 51.7 \\
    
    \midrule
    
    Qwen2VL-7B~\cite{wang2024qwen2vl}    & 80.3 & 70.0 & 61.7 & 48.6 & 56.2 & 40.9 & 20.4 & 54.9 \\
    \rowcolor{mylightblue}\quad\textbf{+\ourmethod}    & 83.0 & 73.5 & 72.4 & 53.1 & 53.6 & 43.6 & 16.5 & \bf{57.3}(\textcolor{red}{$\uparrow$ 2.4\%}) \\
    \bottomrule
\end{tabular}
}
\end{table*}

\section{More Experiments on Other Models.}
\begin{table}[H]
\centering
\caption{\textbf{More Experiments on InternVL3.5-7B and DeepSeek-VL2-Tiny.}}
\label{tab:moremodel}
\footnotesize
\resizebox{0.85\linewidth}{!}{
\begin{tabular}{l ccc}
    \toprule
    Method & StreamingBench& OV-Bench & OVO-Bench  \\
    \midrule    
    InternVL3.5-7B	&75.26	&53.33	&56.49 \\
    \rowcolor{mylightblue}
    \quad\textbf{+\ourmethod} &\textbf{80.02}(\textcolor{red}{$\uparrow$ 4.76\%})&	\textbf{58.78}(\textcolor{red}{$\uparrow$ 5.45\%})	&\textbf{58.81}(\textcolor{red}{$\uparrow$ 2.32\%})\\
    \midrule
    DeepSeek-VL2-Tiny	&53.32	&44.57&	43.23 \\
    \rowcolor{mylightblue}
    \quad\textbf{+\ourmethod}   & \textbf{58.39}(\textcolor{red}{$\uparrow$ 5.07\%})&	\textbf{49.25}(\textcolor{red}{$\uparrow$ 4.68\%})	&\textbf{45.18}(\textcolor{red}{$\uparrow$ 1.95\%})\\
    \bottomrule
\end{tabular}
}
\end{table}

\section{More Visualization Results}
\begin{figure}[H]
  \centering    \centerline{\includegraphics[width=\textwidth]{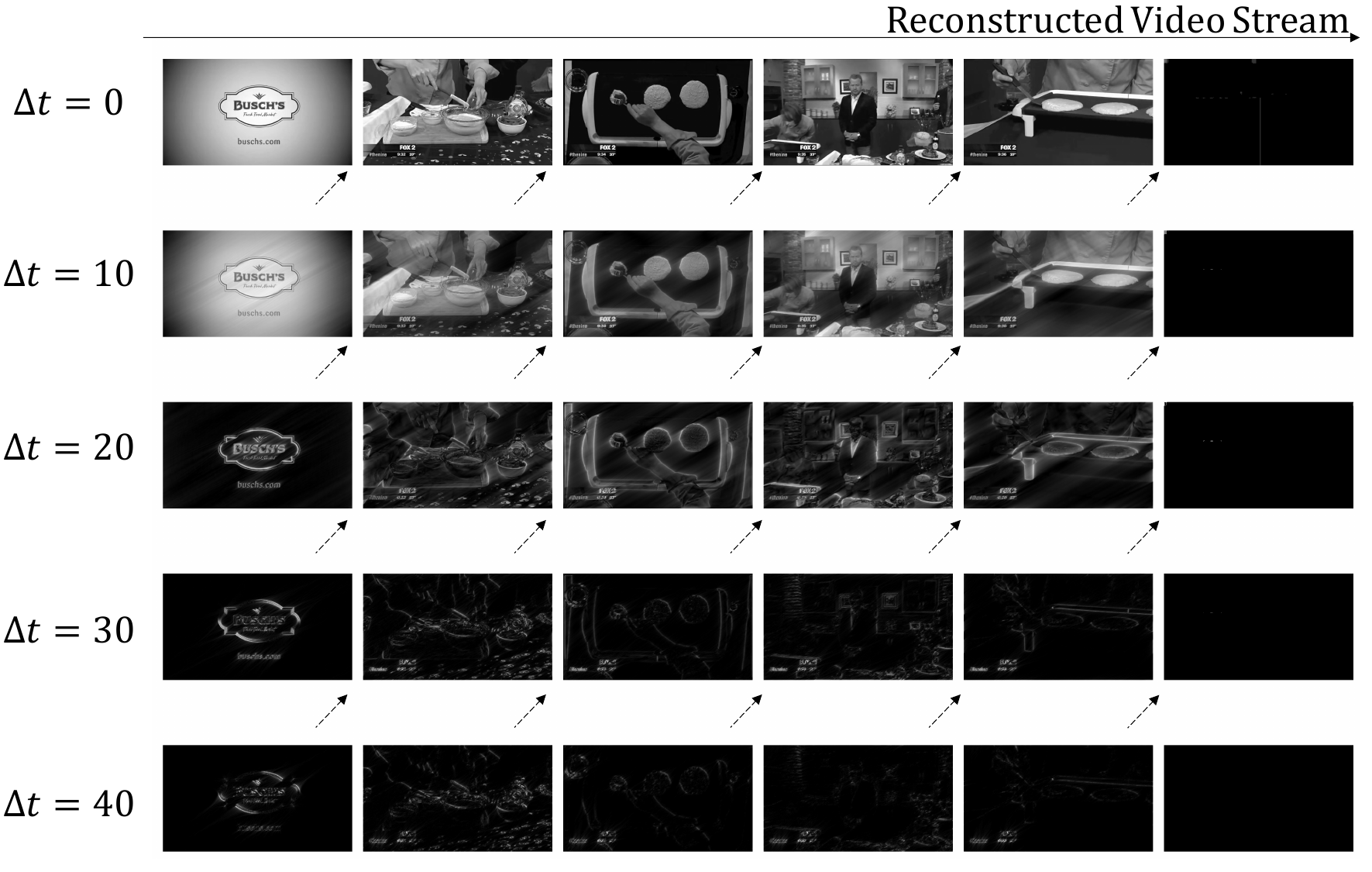}}
    \caption{\textbf{Reconstructed video stream from frequency coefficients.}}
  \label{a1}
\end{figure}

\begin{figure}[H]
  \centering    \centerline{\includegraphics[width=\textwidth]{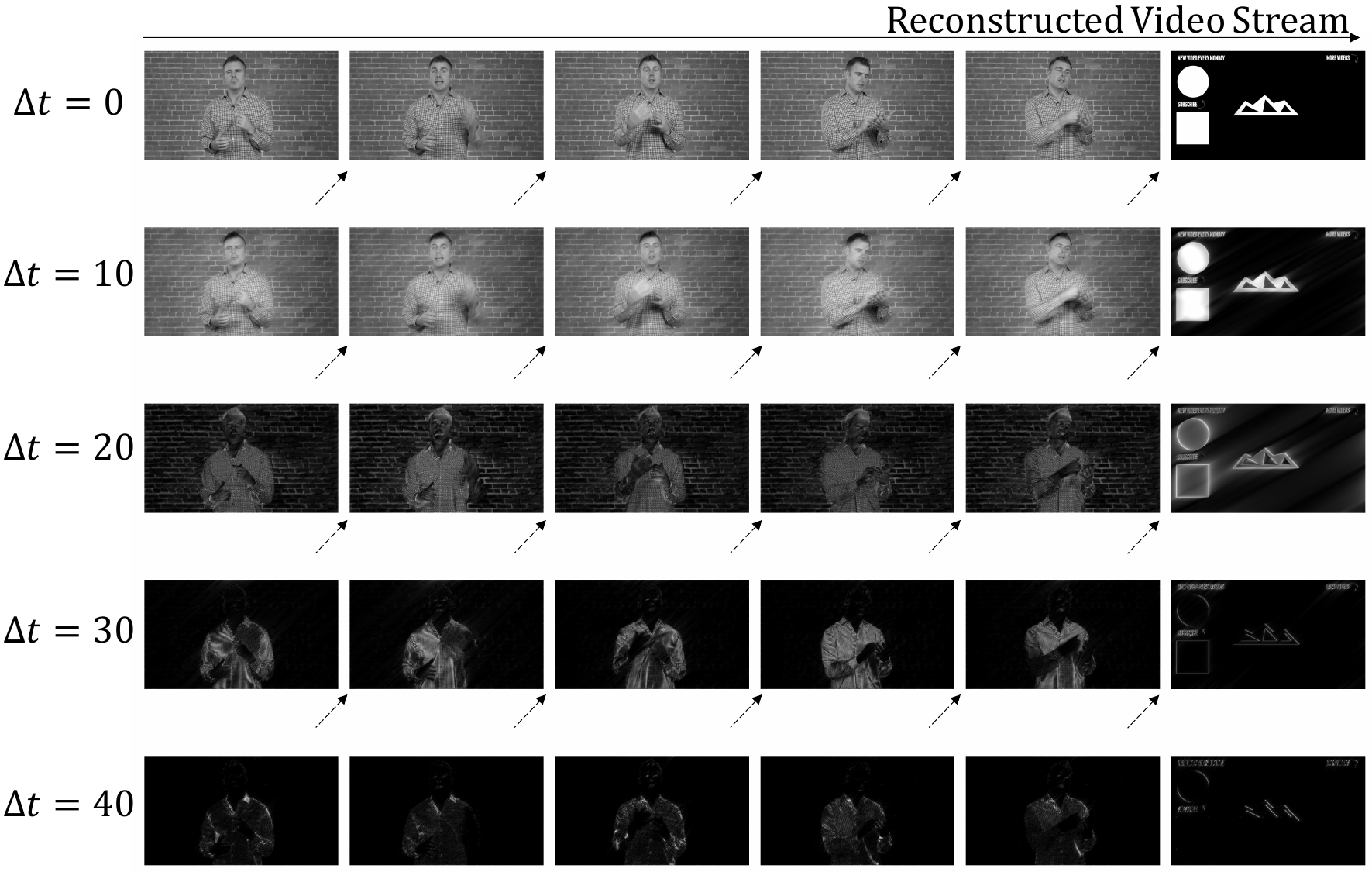}}
    \caption{\textbf{Reconstructed video stream from frequency coefficients.}}
  \label{a2}
\end{figure}

\begin{figure}[H]
  \centering    \centerline{\includegraphics[width=\textwidth]{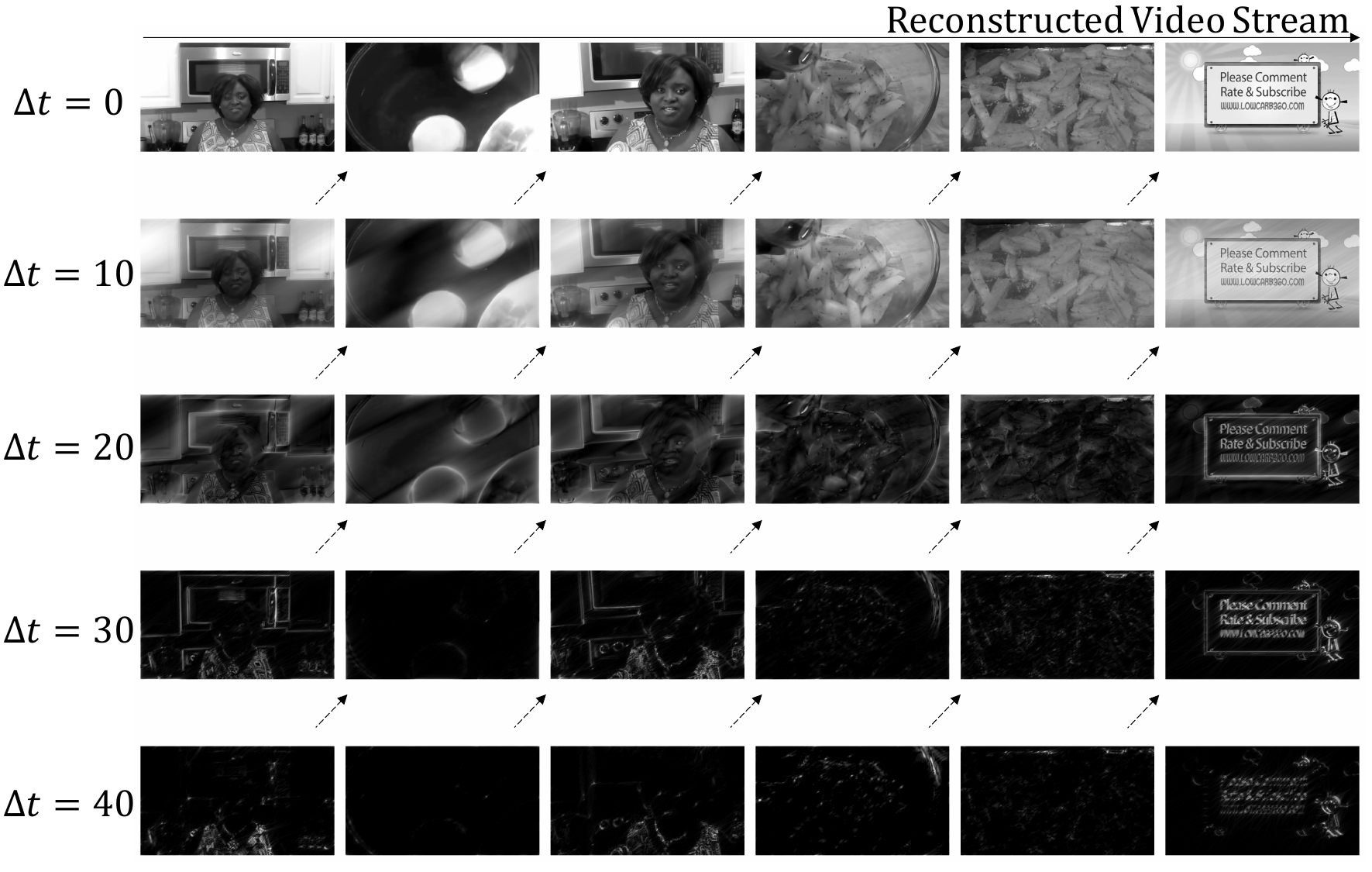}}
    \caption{\textbf{Reconstructed video stream from frequency coefficients.}}
  \label{a3}
\end{figure}

\begin{figure}[H]
  \centering    \centerline{\includegraphics[width=\textwidth]{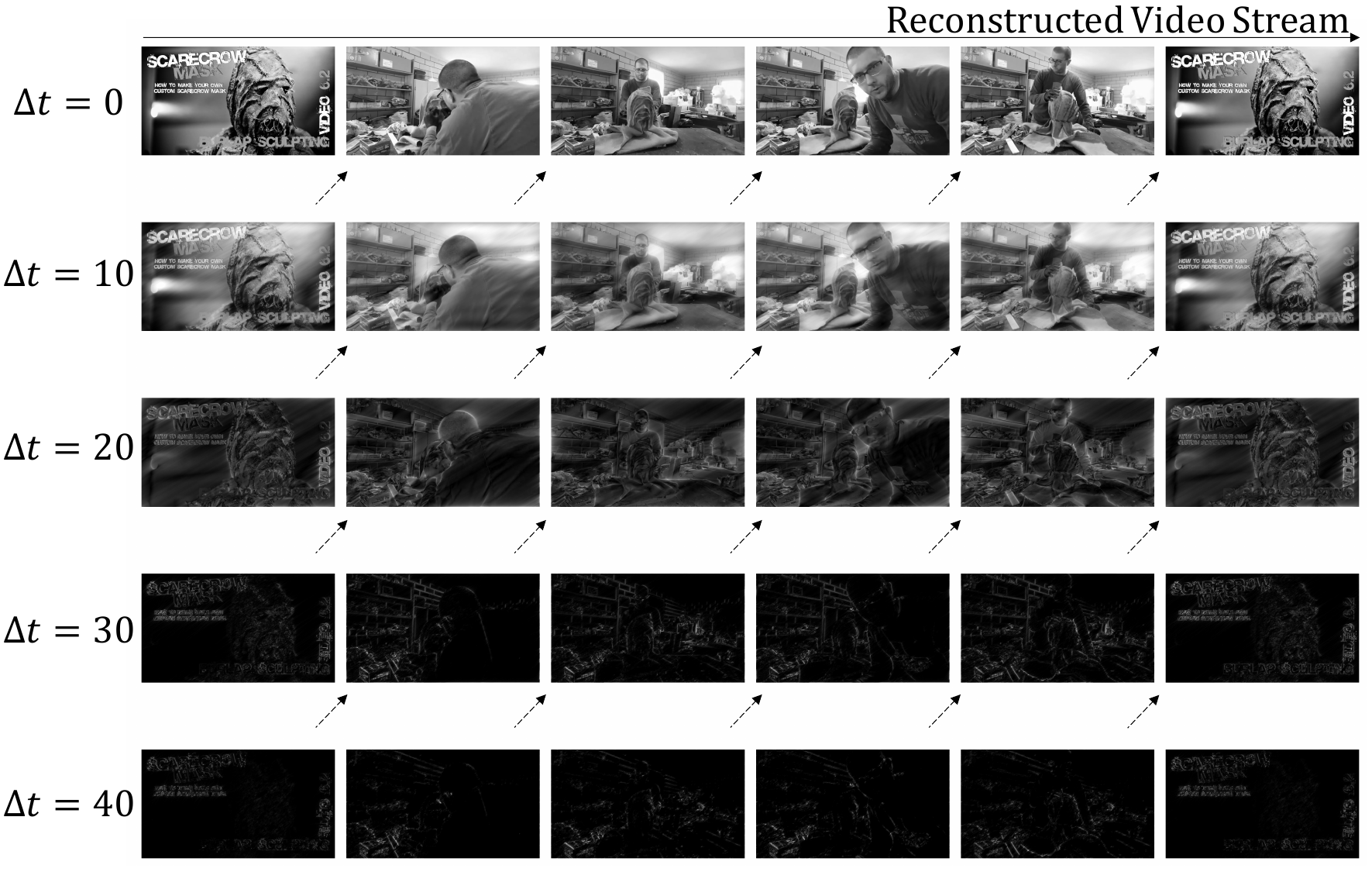}}
    \caption{\textbf{Reconstructed video stream from frequency coefficients.}}
  \label{a4}
\end{figure}

\begin{figure}[H]
  \centering    \centerline{\includegraphics[width=\textwidth]{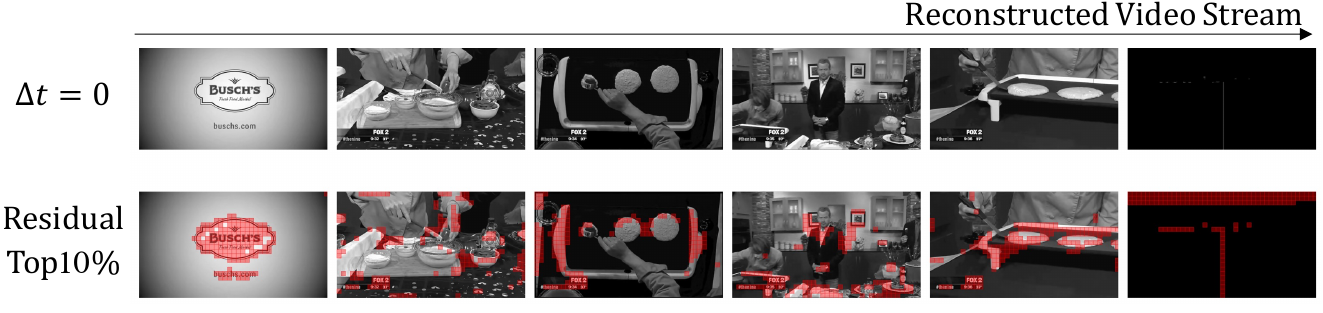}}
    \caption{\textbf{Impact of Salient Residual Tokens.}}
    \label{b1}
\end{figure}

\begin{figure}[H]
  \centering    \centerline{\includegraphics[width=\textwidth]{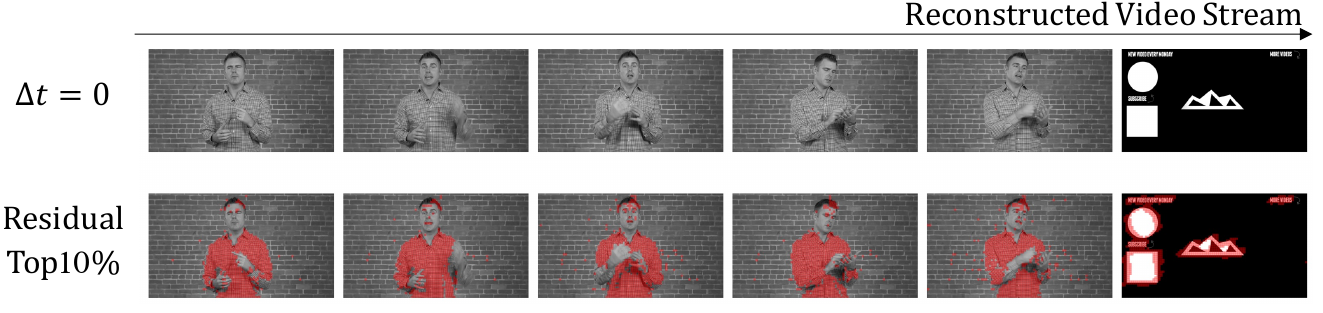}}
    \caption{\textbf{Impact of Salient Residual Tokens.}}
    \label{b2}
\end{figure}

\begin{figure}[H]
  \centering    \centerline{\includegraphics[width=\textwidth]{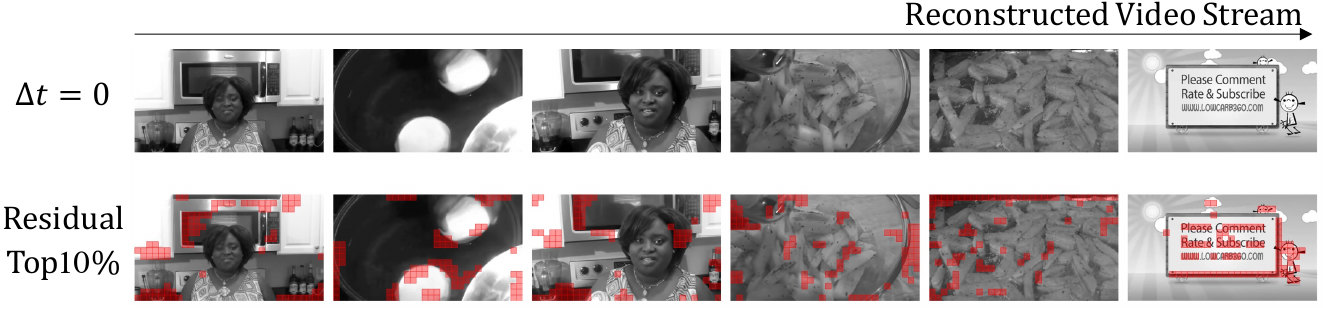}}
    \caption{\textbf{Impact of Salient Residual Tokens.}}
    \label{b3}
\end{figure}

\begin{figure}[H]
  \centering    \centerline{\includegraphics[width=\textwidth]{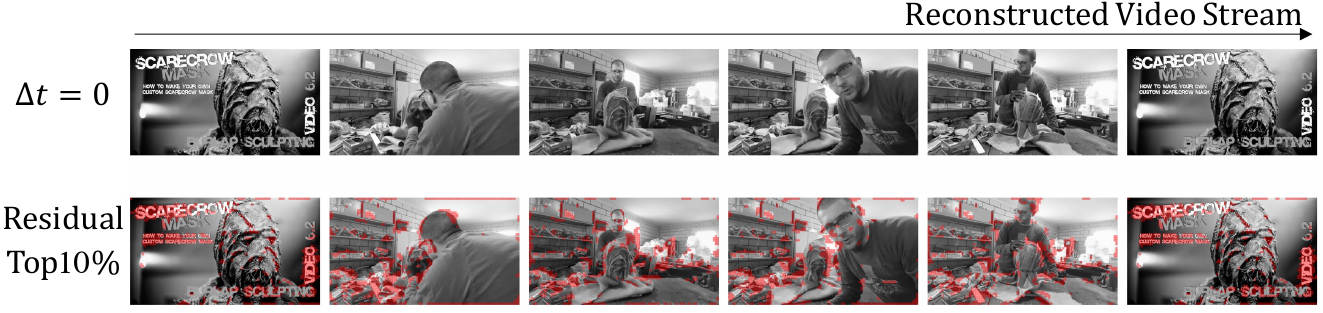}}
    \caption{\textbf{Impact of Salient Residual Tokens.}}
    \label{b4}
\end{figure}

\begin{figure}[H]
  \centering    \centerline{\includegraphics[width=\textwidth]{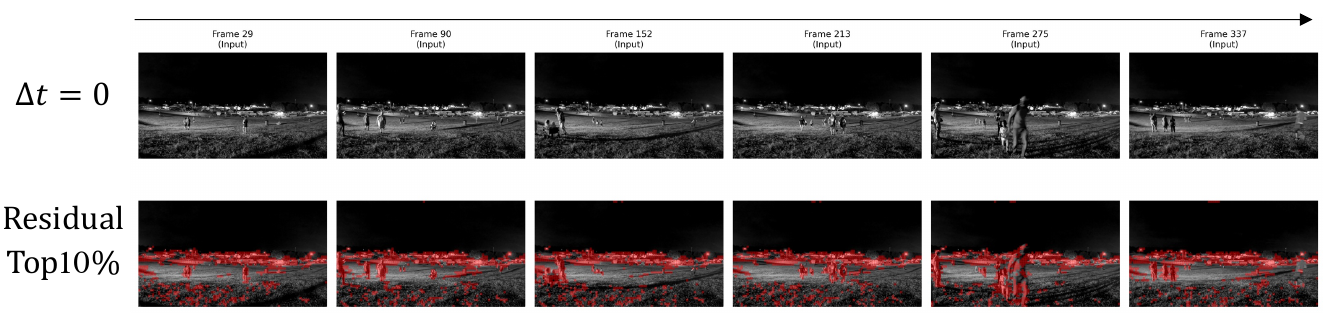}}
    \caption{\textbf{Impact of Salient Residual Tokens.}}
    \label{b5}
\end{figure}

~\cref{a1,a2,a3,a4} show more video frames of the current streaming video reconstructed from the MFM modules based on different DFT coefficients at different steps. We can still observe that MFM is capable of retaining the gist information from the past history, which is consistent with our discussion in the main text.
~\cref{b1,b2,b3,b4,b5} show more residual tokens reconstructed in the MFM module.
From the results, it can be seen that the crucial residual information is basically concentrated in the semantic important areas, which is consistent with our main conclusion in the text.
Importantly, even for video material such as that shown in~\cref{b5}—where the scene is very dark yet full of crucial dialogues or subtle actions—the semantic information can still be effectively captured.
MFM can maintain the historical semantic outline while preserving high-fidelity visual information.


\section{More Case Studies}

\begin{figure*}[h]
  \centering    \centerline{\includegraphics[width=\textwidth]{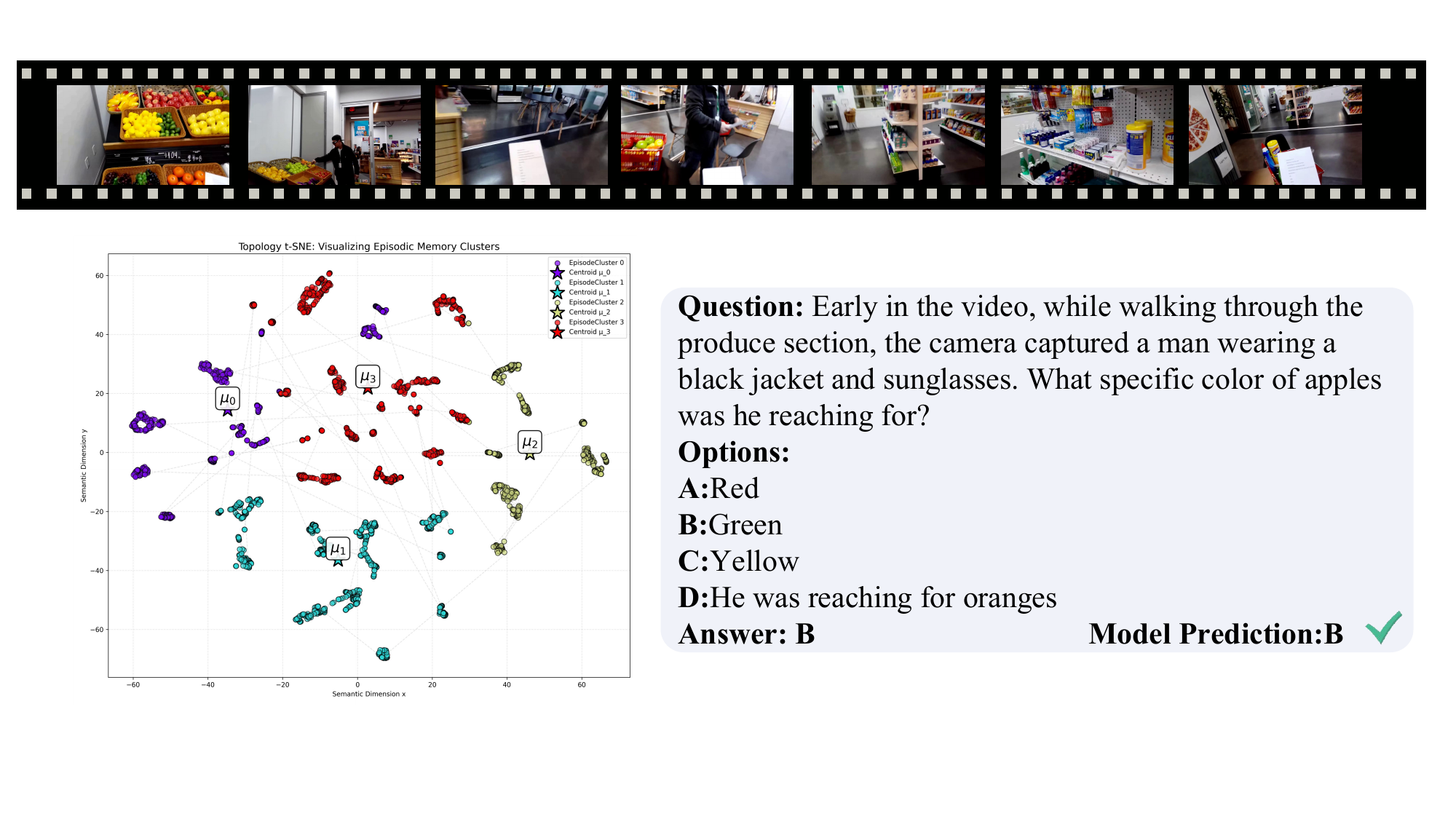}}
    \caption{\textbf{Interpretability analysis of the STM module.}}
    \label{c1}
\end{figure*}

\begin{figure*}[h]
  \centering    \centerline{\includegraphics[width=\textwidth]{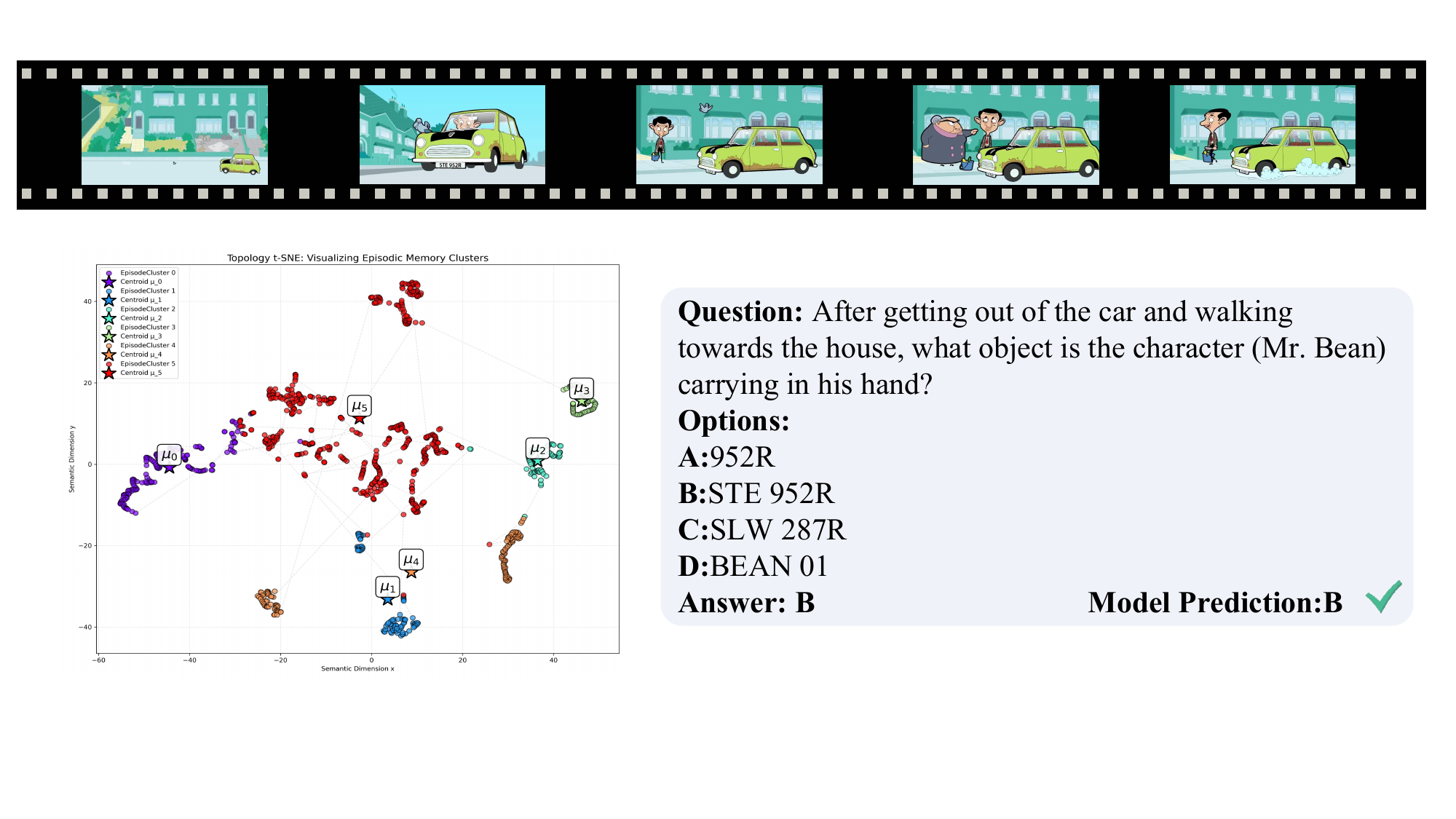}}
    \caption{\textbf{Interpretability analysis of the STM module.}}
    \label{c2}
\end{figure*}

\begin{figure*}[h]
  \centering    \centerline{\includegraphics[width=\textwidth]{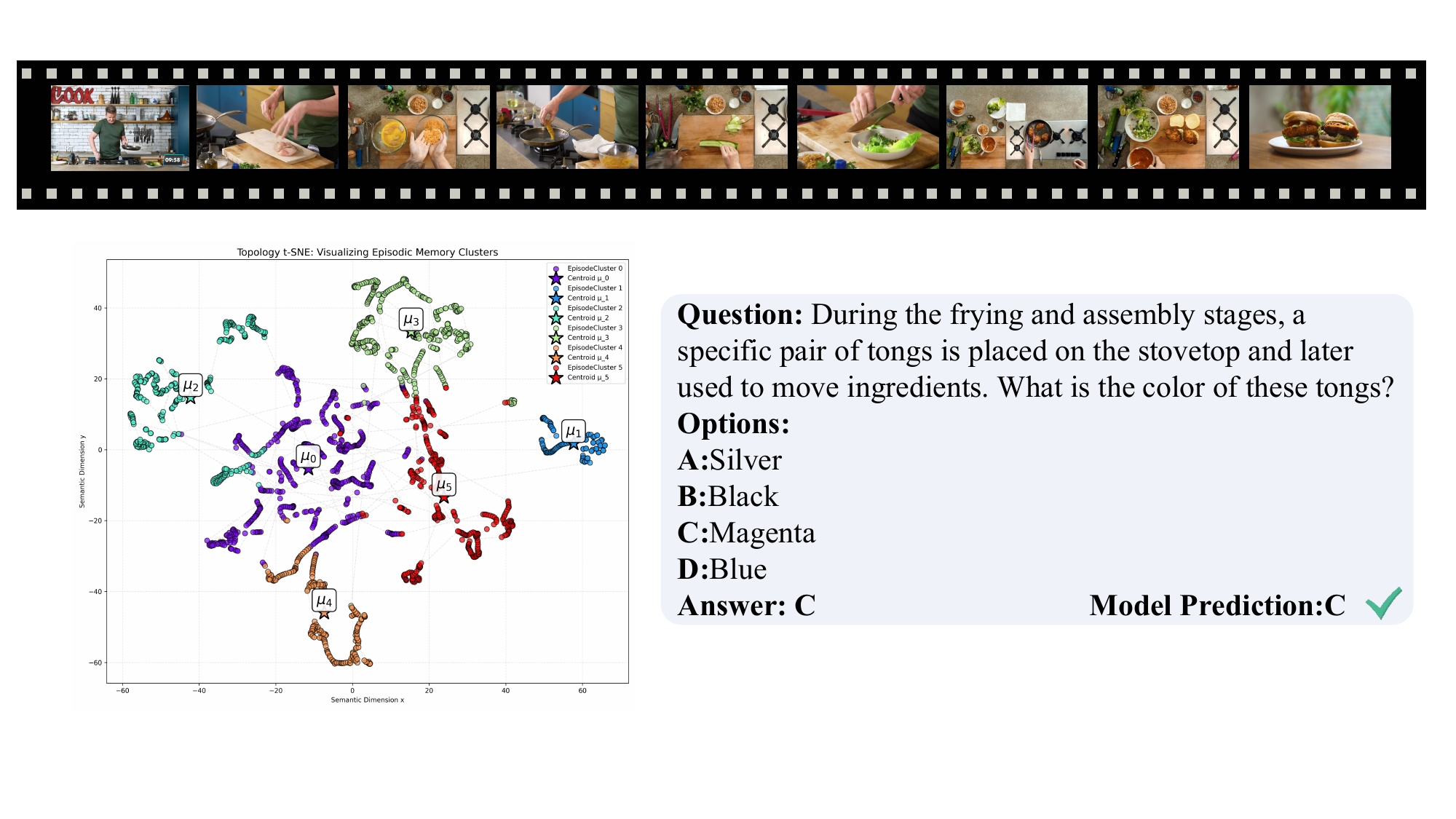}}
    \caption{\textbf{Interpretability analysis of the STM module.}}
    \label{c3}
\end{figure*}


\newpage
\onecolumn
\section*{NeurIPS Paper Checklist}

\begin{enumerate}

\item {\bf Claims}
    \item[] Question: Do the main claims made in the abstract and introduction accurately reflect the paper's contributions and scope?
    \item[] Answer: \answerYes{} 
    \item[] Justification: All major claims stated in the abstract and introduction are thoroughly substantiated by the results and analyses presented in~\cref{Experiments}.

\item {\bf Limitations}
    \item[] Question: Does the paper discuss the limitations of the work performed by the authors?
    \item[] Answer: \answerYes{}{} 
    \item[] Justification: Our work’s limitations are discussed in~\cref{Conclusion}.

\item {\bf Theory assumptions and proofs}
    \item[] Question: For each theoretical result, does the paper provide the full set of assumptions and a complete (and correct) proof?
    \item[] Answer: \answerNA{} 
    \item[] Justification: No theoretical results are included in this paper.

    \item {\bf Experimental result reproducibility}
    \item[] Question: Does the paper fully disclose all the information needed to reproduce the main experimental results of the paper to the extent that it affects the main claims and/or conclusions of the paper (regardless of whether the code and data are provided or not)?
    \item[] Answer: \answerYes{} 
    \item[] Justification: All of the necessary details to make our work reproducible are included in this paper. Our methodology is described in detail in~\cref{Method}, and our experimental setup and implementation details are included in~\cref{Experiments}. Additionally, all of our code and data will be included in the submission and released upon acceptance.

\item {\bf Open access to data and code}
    \item[] Question: Does the paper provide open access to the data and code, with sufficient instructions to faithfully reproduce the main experimental results, as described in supplemental material?
    \item[] Answer: \answerYes{} 
    \item[] Justification: All of the code and data used in this study as well as the necessary documentation to run it will be released upon acceptance.

\item {\bf Experimental setting/details}
    \item[] Question: Does the paper specify all the training and test details (e.g., data splits, hyperparameters, how they were chosen, type of optimizer) necessary to understand the results?
    \item[] Answer: \answerYes{} 
    \item[] Justification: All of the necessary details for testing in terms of experimental setup and implementation details, including training splits and hyperparameter tuning can be found in~\cref{Experiments}.

\item {\bf Experiment statistical significance}
    \item[] Question: Does the paper report error bars suitably and correctly defined or other appropriate information about the statistical significance of the experiments?
    \item[] Answer: \answerYes{} 
    \item[] Justification: Our method is inherently training-free, meaning it does not involve stochastic training processes (e.g., random initialization or data shuffling) that typically require multiple runs to assess variability. Since the approach operates deterministically on fixed pretrained models, identical inputs and configurations, as demonstrated in~\cref{Experiments}, will always produce the same outputs. Results are reproducible across identical hardware and software environments.

\item {\bf Experiments compute resources}
    \item[] Question: For each experiment, does the paper provide sufficient information on the computer resources (type of compute workers, memory, time of execution) needed to reproduce the experiments?
    \item[] Answer: \answerYes{} 
    \item[] Justification: We discuss the local computing resources we utilize in~\cref{Experiments} and detailed time and costs will be put in~\cref{eff}.
    
\item {\bf Code of ethics}
    \item[] Question: Does the research conducted in the paper conform, in every respect, with the NeurIPS Code of Ethics \url{https://neurips.cc/public/EthicsGuidelines}?
    \item[] Answer: \answerYes{} 
    \item[] Justification: We reviewed the NeurIPS Code of Ethics and made sure that our paper conforms to it in every respect.

\item {\bf Broader impacts}
    \item[] Question: Does the paper discuss both potential positive societal impacts and negative societal impacts of the work performed?
    \item[] Answer: \answerYes{} 
    \item[] Justification: Our work promotes interdisciplinary research between AI and neuroscience. This may positively impact future research on interpretable and cognitively aligned AI. As a methodological contribution, it poses no foreseeable negative societal risks.
    
\item {\bf Safeguards}
    \item[] Question: Does the paper describe safeguards that have been put in place for responsible release of data or models that have a high risk for misuse (e.g., pre-trained language models, image generators, or scraped datasets)?
    \item[] Answer: \answerNA{} 
    \item[] Justification: We release no models and the data we release is either already publicly available or purely the output of an LLM doing OpenIE on such data. We believe that this paper poses no such risks.

\item {\bf Licenses for existing assets}
    \item[] Question: Are the creators or original owners of assets (e.g., code, data, models), used in the paper, properly credited and are the license and terms of use explicitly mentioned and properly respected?
    \item[] Answer: \answerYes{} 
    \item[] Justification: We credit the owners of all code, models and data used in this work. Much of this information can be found in~\cref{Experiments}.

\item {\bf New assets}
    \item[] Question: Are new assets introduced in the paper well documented and is the documentation provided alongside the assets?
    \item[] Answer: \answerYes{} 
    \item[] Justification: All of the code and data assets released alongside our paper are appropriately documented for reproducibility.

\item {\bf Crowdsourcing and research with human subjects}
    \item[] Question: For crowdsourcing experiments and research with human subjects, does the paper include the full text of instructions given to participants and screenshots, if applicable, as well as details about compensation (if any)? 
    \item[] Answer: \answerNA{} 
    \item[] Justification: Our paper involves no crowdsourcing or research with human subjects.

\item {\bf Institutional review board (IRB) approvals or equivalent for research with human subjects}
    \item[] Question: Does the paper describe potential risks incurred by study participants, whether such risks were disclosed to the subjects, and whether Institutional Review Board (IRB) approvals (or an equivalent approval/review based on the requirements of your country or institution) were obtained?
    \item[] Answer: \answerNA{} 
    \item[] Justification: Our paper involves no crowdsourcing or research with human subjects.

\item {\bf Declaration of LLM usage}
    \item[] Question: Does the paper describe the usage of LLMs if it is an important, original, or non-standard component of the core methods in this research? Note that if the LLM is used only for writing, editing, or formatting purposes and does \emph{not} impact the core methodology, scientific rigor, or originality of the research, declaration is not required.
    \item[] Answer: \answerNA{} 
    \item[] Justification: The LLMs used, such as ChatGPT, are limited to slight polishing and grammatical corrections of the paper language. They had no role in the design of core research methods, data analysis, experimental procedures, or interpretation of results. Therefore, the use of LLMs did not have an impact on the scientific rigor, originality or substance of the study, so it was not necessary to make a formal declaration in the paper.

\end{enumerate}

\end{document}